\documentclass[conference]{IEEEtran}
\IEEEoverridecommandlockouts

\usepackage{subcaption}
\usepackage{amsmath}
\usepackage{algorithm}
\usepackage{algpseudocode}
\usepackage{listings}
\usepackage{booktabs}
\usepackage{multirow}
\usepackage{url}
\usepackage{verbatim}
\usepackage{zlmtt}
\usepackage{fancyvrb}
\usepackage{subfiles}
\usepackage{enumitem}
\usepackage{color, colortbl}
\usepackage{tikz}
\usepackage[normalem]{ulem}
\usepackage[font=small,labelfont=bf]{caption} 
\usepackage{nicematrix}
\usepackage{tcolorbox}
\usepackage{multicol}
\usepackage{graphicx}
\usepackage{multirow} 
\usepackage{tablefootnote}
\usepackage{threeparttable}
\usepackage{siunitx} 
\usepackage[T1]{fontenc}
\usepackage{hyperref}
\usepackage{cellspace}

\usepackage[framemethod=tikz]{mdframed}

\hypersetup{
    colorlinks=true,
    linkcolor=black,    
    citecolor=black,    
    urlcolor=black,     
    bookmarksnumbered=true,
    bookmarksopen=true
}

\def\BibTeX{{\rm B\kern-.05em{\sc i\kern-.025em b}\kern-.08em
    T\kern-.1667em\lower.7ex\hbox{E}\kern-.125emX}}

\begin{document}

\title{\texttt{DeepCQ}: General-Purpose Deep-Surrogate Framework for Lossy Compression Quality Prediction}

\author{}
\author{\IEEEauthorblockN{1\textsuperscript{st} Khondoker Mirazul Mumenin}
\IEEEauthorblockA{
\textit{University of North Carolina at Charlotte}\\
Charlotte, United States \\
kmumenin@charlotte.edu}
\and
\IEEEauthorblockN{2\textsuperscript{nd} Robert Underwood}
\IEEEauthorblockA{
\textit{Argonne National Laboratory}\\
Lemont, United States \\
runderwood@anl.gov }
\and
\IEEEauthorblockN{3\textsuperscript{rd} Dong Dai}
\IEEEauthorblockA{
\textit{University of Delaware}\\
Newark, United States \\
dai@udel.edu}
\and
\IEEEauthorblockN{4\textsuperscript{th} Jinzhen Wang}
\IEEEauthorblockA{
\textit{University of North Carolina at Charlotte}\\
Charlotte, United States \\
jwang96@charlotte.edu}
\and
\IEEEauthorblockN{5\textsuperscript{th} Sheng Di}
\IEEEauthorblockA{
\textit{Argonne National Laboratory}\\
Lemont, United States \\
sdi1@anl.gov}
\and
\IEEEauthorblockN{6\textsuperscript{th} Zarija Lukić}
\IEEEauthorblockA{
\textit{Lawrence Berkeley National Laboratory}\\
Berkeley, United States \\
zarija@lbl.gov}
\and
\IEEEauthorblockN{7\textsuperscript{th} Franck Cappello}
\IEEEauthorblockA{
\textit{Argonne National Laboratory}\\
Lemont, United States \\
cappello@mcs.anl.gov}
}

\maketitle

\begin{abstract}
  Error-bounded lossy compression techniques have become vital for scientific data management and analytics, given the ever-increasing volume of data generated by modern scientific simulations and instruments. Nevertheless, assessing data quality post-compression remains computationally expensive due to the intensive nature of metric calculations. 
  In this work, we present a general-purpose deep-surrogate framework for lossy compression quality prediction (DeepCQ), with the following key contributions: 1) We develop a surrogate model for compression quality prediction that is generalizable to different error-bounded lossy compressors, quality metrics, and input datasets; 2) We adopt a novel two-stage design that decouples the computationally expensive feature-extraction stage from the light-weight metrics prediction, enabling efficient training and modular inference; 3) We optimize the model performance on time-evolving data using a mixture-of-experts design. Such a design enhances the robustness when predicting across simulation timesteps, especially when the training and test data exhibit significant variation.
  We validate the effectiveness of DeepCQ on four real-world scientific applications. Our results highlight the framework's exceptional predictive accuracy, with prediction errors generally under 10\% across most settings, significantly outperforming existing methods. Our framework empowers scientific users to make informed decisions about data compression based on their preferred data quality, thereby significantly reducing I/O and computational overhead in scientific data analysis.
\end{abstract}


\section{Introduction} \label{sec:intro}

With the advent of exascale High-Performance Computing (HPC) systems, high-fidelity and high-resolution scientific simulations are generating unprecedented volumes of data, typically represented as multi-dimensional tensors, posing significant challenges for storage, transmission, and analysis. For instance, the Climate Earth Science Model (CESM)~\cite{Hurrell2013} simulates Earth's climate over various periods and can produce several petabytes of data in a short time. Specifically, CESM for the Coupled Model Intercomparison Project generates approximately 2.5 PB of data, with around 170 TB refined and submitted to the Earth System Grid~\cite{Cinquini2014}.
Similarly, a Nyx simulation with a resolution of $4096^3$ (i.e., $0.5\times2048^3$ mesh points in the coarse level and $0.5\times4096^3$ in the fine level) can produce up to 1.8 TB per snapshot, requiring a total of 1.8 PB of storage for five runs with 200 snapshots each~\cite{wang2023tac}. Despite improved simulation fidelity, the sheer volume of data often leads to the loss of critical physics during subsequent analysis.

Error-bounded lossy compression techniques~\cite{di2016fast,Lindstrom2014,MGARD2,tao2017significantly,liang2018error,zhao_significantly_2020,zhao2021optimizing,liang2021error,TTHRESH,burtscher2009fpc} have emerged as essential tools for domain scientists to manage this data deluge. These techniques reduce data volume while ensuring that the numerical deviation between original and decompressed data remains within a user-defined error tolerance. By trading accuracy for reduced storage requirements, these compressors improve performance in tasks such as parallel I/O on shared filesystems and scientific visualization. Consequently, many lossy compression techniques have been integrated into modern scientific data management middleware~\cite{adios09,10046061}.

However, the effectiveness of lossy compressors heavily depends on the user-defined error bound, which caps the maximum absolute error between the original and decompressed data. While this ensures strict adherence to numerical constraints, it often fails to capture more application-relevant quality metrics, such as Peak Signal-to-Noise Ratio (PSNR) and Structural Similarity Index Measure (SSIM)~\cite{Wang:2004:ssim}, and its variants, such as dSSIM~\cite {Baker2022}. These metrics are frequently more meaningful to scientists but are not directly correlated with the error bound. For example, PSNR is derived from maximum data values and mean squared error (MSE), while SSIM evaluates structural similarity between datasets. 

This disconnection between what scientists prioritize (various quality metrics) and what they can control (error bounds) complicates the effective use of lossy compressors. Relationships between error bounds and quality metrics are often non-linear and highly variable across applications~\cite{underwood_optzconfig_2022}.  As a result, scientists must resort to trial-and-error processes involving repeated cycles of selecting an error bound, performing compression and decompression, and calculating quality metrics until satisfactory results are achieved \cite{underwood_optzconfig_2022}, incurring significant time and resource costs.

\begin{figure}[t]
    \centering
    \includegraphics[width=\columnwidth]{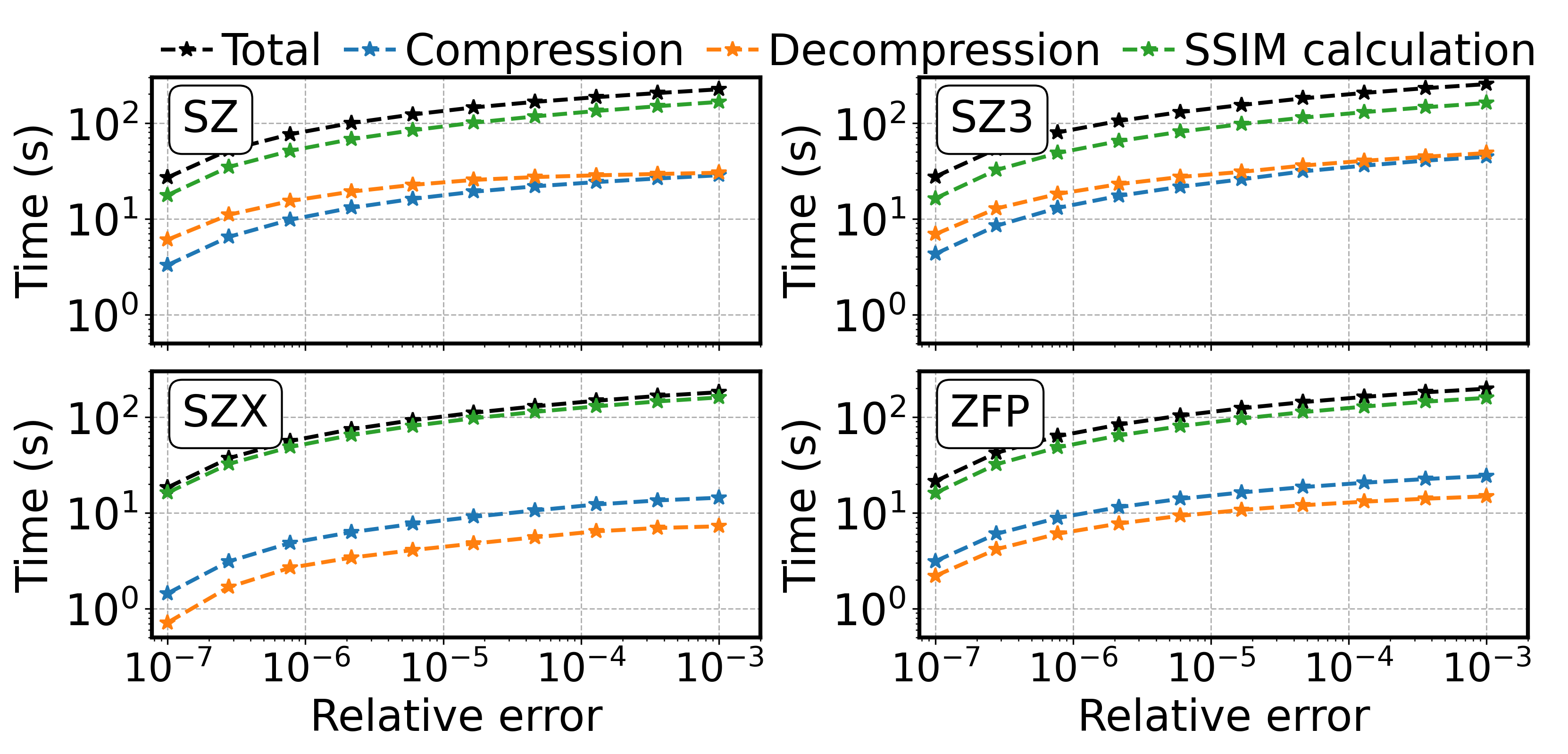}
    \vspace{-16pt}
    \caption{The cumulative time to examine data quality across 10 error bounds. Dataset: Dark matter density; dimension:$[512 \times 512 \times 512]$.}
    \vspace{-16pt}
    \label{fig:1}
\end{figure}

In Fig.~\ref{fig:1}, we illustrate the cumulative time required to explore 10 error bounds for four widely used lossy compressors. Due to the overhead of compression, decompression, and metrics calculation, evaluating a single metric (SSIM) across merely 10 error bounds can take up to 250 seconds (SZ3). The overall evaluation time scales further with the number of compressors, quality metrics, and error bounds considered.


As new compressors with novel principles of compression (prediction, transform, deep learning, etc.) are being introduced, applications may benefit differently from various compressor designs, further exacerbating this problem, given a large search space. The size of this search space poses a significant challenge for achieving peak performance, especially for non-expert users who lack an intuitive understanding of which compressors are well-suited to particular problems and must resort to search-based methods \cite{underwood_optzconfig_2022}.

Motivated by excessive overhead in compression quality calculation and the challenges of navigating through the ample space of exploration in compressor selection and configuration, we propose \textbf{\texttt{DeepCQ}}, a general-purpose deep-surrogate framework, for compression quality prediction. DeepCQ efficiently predicts multiple compression quality metrics for input data and error bounds across various lossy compressors. The framework is scalable and extensible, allowing seamless integration of future compressors and additional quality metrics, thereby distinguishing it from prior approaches that are handcrafted for limited compressors and quality metrics~\cite{tao2019optimizing,jin2022improving,khan_secre_2023}.

Our key contributions are summarized as follows:
\begin{itemize}
	\item We successfully design a general-purpose deep surrogate model that can accurately predict various compression quality metrics for various lossy compressors and input scientific data,
    with the mean absolute percentage error {\underline{less than 10\% in 272 out of 285 cases}}.
    \item Inspired by transfer learning techniques, we propose a novel two-stage design that decouples the expensive feature extraction from the actual quality prediction. Such a design enables feature extraction models to be shared across data fields within the same scientific application, thereby significantly reducing training overhead. It also enables modularity at inference time, allowing users to deploy customized metric predictions for efficiency.
	\item We further enhance the robustness of metrics prediction, especially across data with varying compression quality, via a mixture-of-expert design. Such a design improves prediction accuracy on applications with limited training samples but high compression quality variation.
\end{itemize}

To the best of our knowledge, DeepCQ is the first-of-its-kind surrogate-based compression quality prediction framework that delivers accurate predictions effectively for multiple quality metrics across various lossy compressors. By enabling fast compressor selection and configuration, DeepCQ will empower scientific users to make informed decisions about data compression based on their preferred data quality, leading to better balancing across storage cost, fidelity, and computational overhead in exascale scientific workflows. 

The remainder of the paper is organized as follows. Section~\ref{sec:related} discusses the related work on error-bounded lossy compressors, and compression ratio and quality prediction.
In Section~\ref{sec:design}, we detail the design of DeepCQ, with an emphasis on the two-stage deep surrogate model design and mixture-of-expert enhancement. In Section~\ref{sec:eval}, we assess the performance of DeepCQ and compare it with existing compression quality prediction approaches, followed by ablation studies.
Section~\ref{sec:discussion} outlines the limitations of this work and provides directions for future research, followed by conclusions in Section~\ref{sec:conclusion}.

\section{Related Work and Motivation} \label{sec:related}

\subsection{Error-bounded Lossy Compression}

Error-bounded lossy compression techniques have been the predominant focus for data reduction, as they effectively reduce the volume of scientific data while respecting the users' accuracy requirements.
Based on the way of data decorrelation, lossy compression can generally be divided into \textit{prediction-based} and \textit{transform-based}.

Prediction-based lossy compressors essentially achieve compression by removing data redundancy using predictive models.
SZ2~\cite{di2016fast,tao2017significantly,liang2018error,zhao_significantly_2020}, an example of prediction-based compression techniques, starts by predicting each data point using a curve-fitting model based on the neighboring data point. Then, it performs quantization on the difference between the original and predicted value, followed by applying Huffman coding and lossless compression techniques to reduce the bit rate of each quantization code. Its successor, SZ3~\cite{SZ3,Zhao2021,liu2024high}, further enhances the compression performance by introducing a modular design with interchangeable components (predictors, quantization methods, encoders, and lossless backends) so that the compressor can be tailored toward input datasets and achieve the best-fit compression ratio.
SZx~\cite{SZx} is another SZ family variant that primarily focuses on providing ultra-fast compression speed. It is designed to use lightweight operations and is optimized for CPU and GPU architectures while preserving relatively high compression ratios.

On the other hand, transform-based lossy compressors usually adopt data transform schemes to remove data redundancy. Then, various encoding methods are used to represent the transform coefficients. 
For example, ZFP~\cite{Lindstrom2014} first adopts a non-orthogonal transform on the data within each block, then a customized variable-length embedded encoding to represent transform coefficients one bit-plane at a time. Due to the block-wise design, ZFP offers superior scalable performance and random access to compressed data blocks.
{SPERR~\cite{sperr} similarly adopts a biorthogonal wavelet transform on chunks of input data. Then it adopts a SPECK~\cite{pearlman2004efficient,tang2006three}-an inspired scheme to encode the outliers efficiently. SPERR also offers scalable execution capability as well as point-wise error tolerance, which is often desired by many scientific applications.}

\subsection{Compression Ratio and Quality Prediction}
In practice, it is critical for users to understand how a lossy compressor with a given error bound impacts the final compression ratio and the quality of the resulting compressed data. As described, this is a non-trivial task. Many efforts have been made to address this challenge. We discuss existing approaches and their limitations, motivating our work.

\subsubsection{Compression Ratio Prediction}
First, users need to understand how much space can be saved with certain lossy compressors and error bounds. 
Existing attempts mainly achieve this in two steps: \textit{1) extracting meaningful features from input data that correlate well with compression ratio} and \textit{2) making accurate predictions based on the extracted features}.

The attempts on data feature extraction can be further categorized into two strategies: \textit{1) autonomous extracting from down-sampled data} and \textit{2) assembling statistical data features by hand}.
{Lu~\cite{8425188}, Wang~\cite{wang2019compression, wang2023zperf}, Tao~\cite{tao2019optimizing} and Khan~\cite{khan_secre_2023} fall into down-sampling-based category. They also investigated the impact of different sampling methods and concluded that the prediction could be more accurate if their sampling method matches the way compressors process input data~\cite{8425188,khan_secre_2023}. For example, as ZFP performs local transform and bitplane encoding in $4^d$ blocks, sampling at the same size can yield unbiased compression ratio prediction.

{On the other hand, Qin~\cite{qin2020estimating}, Underwood~\cite{underwood2023black}, Krasowska~\cite{krasowska2021exploring} and Ganguli~\cite{10319944} hand-crafted the data features. The early work by Qin~\cite{qin2020estimating} relied on simple data features (mean, variance, standard deviation, quartiles, histograms) and compressor-specific metrics (curve-fitting hit ratio, Huffman tree size, etc.) to help predict the compression ratio. More recent work like Underwood~\cite{underwood2023black} designed truncated SVD and quantized entropy as predictors for compression ratio and achieved high prediction accuracy. Krasowska~\cite{krasowska2021exploring} further showcased on 2D datasets that data correlation could be a good feature in compression ratio prediction. Ganguli~\cite{10319944} further proposed hand-picked data features (Spatial Diversity, Spatial Correlation, Generic Distortion Measurement, Coding Gain, and Spatial Smoothness) that have been studied to have a strong correlation with data compressibility~\cite{underwood2023black}.}

After extracting the data features, the next step is to make accurate predictions. Lu~\cite{8425188}, Tao~\cite{tao2019optimizing}, Wang~\cite{wang2019compression, wang2023zperf} and Khan~\cite{khan_secre_2023} explored statistical models, based on the empirical study on the correlation between compression ratio and multiple compression metrics (prediction hit ratio, Huffman tree size, distribution of SZ prediction error). 
Meanwhile, the work by Qin~\cite{qin2020estimating}, Underwood et al.~\cite{underwood2023black} and
Ganguli~\cite{10319944} adopted data-driven approaches like regression models or Deep Neural Network (DNN) models for compression ratio prediction.

\subsubsection{Compression Quality Prediction}
Recently, as the quality of the compressed data has become increasingly critical for domain scientists, a few new studies have emerged in predicting data quality-relevant metrics, such as SSIM or PSNR, in addition to compression ratio. 
The strategies adopted, however, are not fundamentally different. For example,
Jin's work~\cite{jin2022improving} models compression ratio, PSNR, and SSIM via a mathematical model specially designed for prediction-based compressors. Such a model also leverages the down-sampled data.
A recent work by Mumenin et al.~\cite{mumenin2024qualitynet} also adopts a deep learning-based approach that takes original scientific data and error bound values as input and learns to predict compression quality like PSNR, SSIM, and compression error autocorrelation (AC). It is among the first to learn data features using a DNN model. 

{To summarize, predicting the compression ratio and even data quality of lossy compressors has been extensively studied. However, these efforts are all heavily crafted for a particular compressor or a given target (compression ratio prediction). Applying them to a different compressor or even a new version of the compressor in the same family incurs either a complete redesign of the method (hand-picked feature-based) or a retraining of the entire machine learning model. Either way yields significant costs. It becomes notoriously difficult for users to compare different compressors and select the best compressor for their data.}

\subsection{Challenges and Motivation}

Among existing studies, the work by Mumenin et al.~\cite{mumenin2024qualitynet} is the closest to supporting simultaneous prediction of compression ratio and data quality. However, their approach requires \textit{retraining the entire DNN model on each dataset for every compressor and quality metric}, resulting in excessive computational cost and a substantial storage footprint for pre-trained models. For instance, when applied to the Nyx cosmology simulation, predicting the compression ratio, PSNR, and SSIM across four data fields and five lossy compressors requires training the model 60 times, introducing significant overhead and storage demands. This inefficiency renders the approach impractical for large-scale scientific applications, which often involve hundreds of data fields and numerous quality metrics across multiple compressors~\cite{underwood_optzconfig_2022,pinard_assessing_2020}. Furthermore, their method demonstrated suboptimal performance on the 3D Hurricane Isabel dataset, requiring a 3D-to-2D data conversion to achieve usable results.

Motivated by the substantial training and storage costs associated with maintaining separate datasets for retraining DNN models, this study proposes a novel and general-purpose approach. Instead of training a large model to predict each quality metric directly from raw data and error bounds, we separate the costly feature extraction process from the metric prediction, allowing each network to be trained and evaluated independently, thereby improving the efficiency during both training and inference.

The framework further generalizes data feature extraction, where a shared feature extraction network can be reused across data fields within the same application. This design eliminates the need to store multiple pre-trained feature extractors and allows the trained extractor to be shared with the research community for easy adaptation to new quality metrics or quantities of interest. 

Additionally, the framework adopts a modular inference strategy that is inherently extensible, supporting seamless integration of new compressors and quality metrics. Each compressor–metric pair operates within its own independent prediction module (or “head”), enabling users to select only the components relevant to their tasks. For example, a climate scientist may deploy prediction heads for SSIM and temperature variation while excluding unrelated metrics. By extracting reusable features in a one-time preprocessing step, the framework avoids repeated feature extraction and enables parallel predictions across multiple compressors and metrics, thereby achieving high scalability and flexibility.

\begin{table}[]
\caption{Comparison with existing approaches}
\setlength{\tabcolsep}{2pt}
\renewcommand{\arraystretch}{1.45}
\vspace{-5pt}
\centering
\label{tab:position}
\resizebox{\columnwidth}{!}{%
\begin{tabular}{ccccc}
\hline \hline
\textbf{Methods}                  & \textbf{Feature Type}                                        & \textbf{Model Type}                                              & \textbf{Compressor}                                                     & \textbf{Metrics}                                          \\ \hline
Tao2019~\cite{tao2019optimizing}  & Sampling                                                     & Statistical                                                      & SZ2, ZFP                                                                & CR, PSNR                                                  \\ \hline
Jin2022~\cite{jin2022improving}   & Sampling                                                     & Analytical                                                       & SZ3                                                                     & \begin{tabular}[c]{@{}c@{}}CR, PSNR,\\  SSIM\end{tabular} \\ \hline
Khan2023~\cite{khan_secre_2023}   & Sampling                                                     & \begin{tabular}[c]{@{}c@{}}Handcrafted \\ surrogate\end{tabular} & \begin{tabular}[c]{@{}c@{}}SZ3, SZx,\\  ZFP, SPERR\end{tabular}         & CR                                                        \\ \hline
DeepCQ                            & \begin{tabular}[c]{@{}c@{}}Learned \\ embedding\end{tabular} & \begin{tabular}[c]{@{}c@{}}DNN \\ surrogate\end{tabular}         & \begin{tabular}[c]{@{}c@{}}SZ2, SZ3, \\ SZx, ZFP, SPERR\end{tabular} & \begin{tabular}[c]{@{}c@{}}CR, PSNR, \\ SSIM\end{tabular}    \\ 
\hline \hline
\end{tabular}%
}
\vspace{-15pt}
\end{table}

We position the design of DeepCQ relative to existing compression quality prediction approaches in Table~\ref{tab:position}. As a general-purpose framework, DeepCQ offers the most comprehensive support for diverse compressors and quality metrics while maintaining effortless extensibility, making it a practical solution for real-world scientific applications. Through its innovative design, DeepCQ bridges the gap between compression error bounds and scientifically meaningful quality metrics or quantities of interest. Its efficient prediction capability enables domain scientists to make informed decisions on compression configurations without relying on exhaustive trial-and-error experimentation. For example, researchers can optimize compression settings to preserve structural similarity or critical scientific features while minimizing computational cost. The following section presents the detailed design and implementation of the proposed framework.

\section{Deep-surrogate framework for compression quality prediction} \label{sec:design}

In this section, we begin by discussing the problem formulation and then provide the details of the framework design. But first, we list all the key terms with descriptions in Table~\ref{tab:term}. 

\subsection{Problem Formulation}

Compression quality metrics play a crucial role in assessing the impact of lossy compression on data. By referencing the original dataset, full-reference quality metrics~\cite{pedersen2012full} quantify the distortion introduced by lossy compression, making them invaluable for users to assess the quality of compressed data.

Typically, compression quality metrics are \underline{scalar values} derived from the differences between the original and decompressed data. These metrics provide an efficient summary of compression errors, especially when dealing with \underline{multi-dimensional tensors} where examining each data point individually is impractical. Some metrics, such as PSNR and MSE, capture statistical information about point-wise compression errors. Others, like SSIM, focus on local features and perceptual quality. These metrics respond to the compression error with varying levels of sensitivity.

Considering the expensive calculation required in obtaining these compression quality metrics, our goal is to learn a mapping relationship between error bounds and compression quality metrics for approximation. However, this task is complicated by the diversity of algorithms and designs employed by different lossy compressors. Even under identical error bounds, the effects of compression on data quality can vary significantly across compressors~\cite{10025484}. Additionally, the characteristics of the input data itself also influence how compression affects data quality, further complicating this relationship.

Given this interplay between the compression algorithm, input data, and error bounds, capturing compression quality is inherently complex. An explicit mapping from error bounds to compression quality cannot be directly derived for most quality metrics, and therefore, we resort to black-box deep neural network (DNN) models as a surrogate, with a theory foundation provided by Hornik~\cite{HORNIK1989359} and practical validation by Mumenin's work~\cite{mumenin2024qualitynet}. However, due to the significant inefficiency, it is not practical to apply Mumenin's work toward compression quality prediction widely on various compression quality metrics, lossy compressors and datasets.

\begin{table}[t]
\setlength{\tabcolsep}{20pt}
\caption{Key terminologies used in this paper}
\renewcommand{\arraystretch}{1.1}
\vspace{-5pt}
\centering
\scriptsize
\label{tab:term}
\resizebox{\columnwidth}{!}{%
\begin{tabular}{ll}
\hline \hline
\textbf{Terms}& \textbf{Description}                   \\ \hline
PSNR          & Peak Signal-to-Noise Ratio             \\ \hline
SSIM          & Structural Similarity Index Metric     \\ \hline
CR            & Compression Ratio                      \\ \hline
PE            & Percentage Error                       \\ \hline
MAPE          & Mean Absolute Percentage Error         \\ \hline
DFE-NN        & Data Feature Extraction Network        \\ \hline
EFE-NN        & Error bound Feature Extraction Network \\ \hline
Pred-NN       & Prediction Network                     \\ \hline
MoE           & Mixture of Expert                      \\ 
\hline \hline
\end{tabular}%
}
\vspace{-15pt}
\end{table}

Our objective is to substantially enhance the efficiency of deep-surrogate-based compression quality prediction methods, enabling their adoption as a general-purpose solution across diverse quality metrics, lossy compressors, and datasets. In this work, we focus on three representative and broadly applicable compression quality metrics as prediction targets: PSNR, SSIM, and Compression Ratio (CR)\footnote{We extend the definition of compression quality metrics to include compression ratio, as it is an essential measurement for assessing data reduction performance in lossy compression.}. Specifically, PSNR and SSIM are widely adopted to quantify data distortion introduced during compression, while CR provides critical insight into the achievable data reduction efficiency.

\subsection{General Principle: A Two-Stage Approach}

The primary inefficiency in Mumenin’s approach~\cite{mumenin2024qualitynet} arises from the need to retrain the entire model from scratch for each dataset. Consequently, the overhead associated with model training, management, and storage scales linearly with the number of datasets, rendering the method impractical for large-scale adoption. However, data fields within the same scientific application often exhibit similar spatial or temporal characteristics. As illustrated in Fig.~\ref{fig:2}, multiple data fields in the Hurricane application share common patterns, such as hurricane formation structures. Therefore, retraining the entire surrogate model for each field is redundant, as these shared data characteristics can be effectively captured through a unified feature extraction backbone.

\begin{figure}[t]
  \centering
  \begin{subfigure}[t]{0.24\columnwidth}
    \centering
    \includegraphics[width=\linewidth]{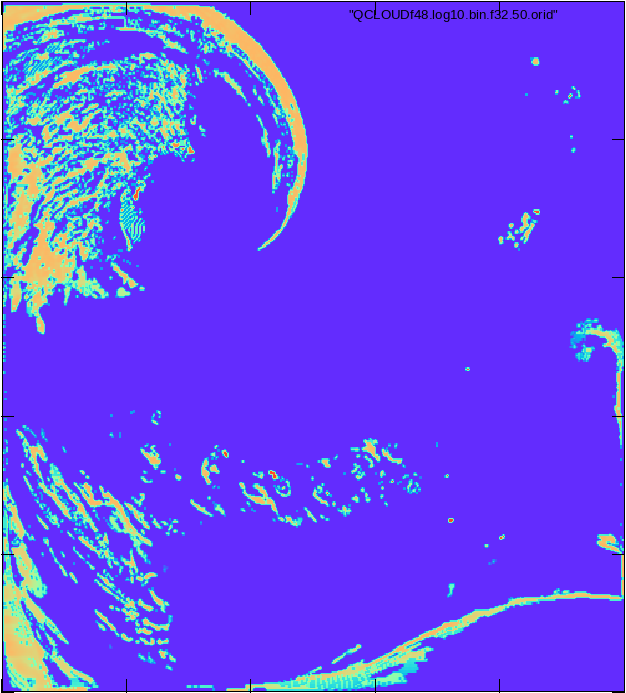}
    \caption{QCLOUD}
    \label{fig:a}
  \end{subfigure}
  \hfill
  \begin{subfigure}[t]{0.24\columnwidth}
    \centering
    \includegraphics[width=\linewidth]{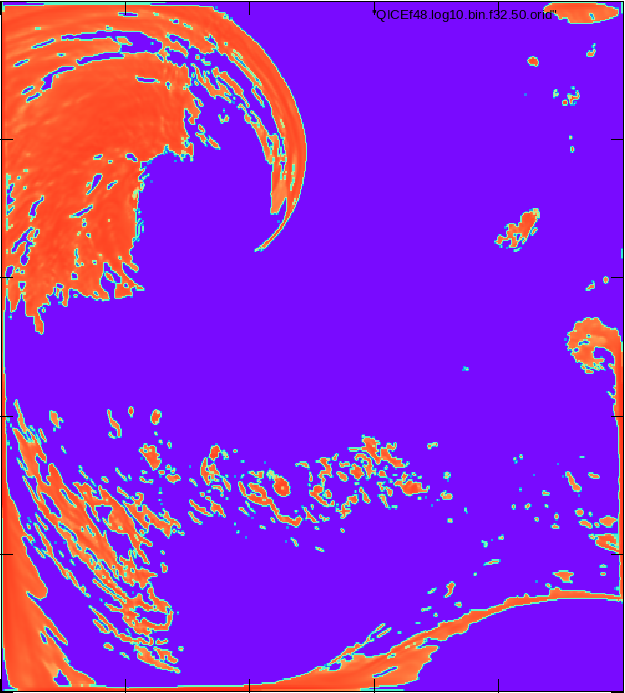}
    \caption{QICE}
    \label{fig:d}
  \end{subfigure}
  \hfill
  \begin{subfigure}[t]{0.24\columnwidth}
    \centering
    \includegraphics[width=\linewidth]{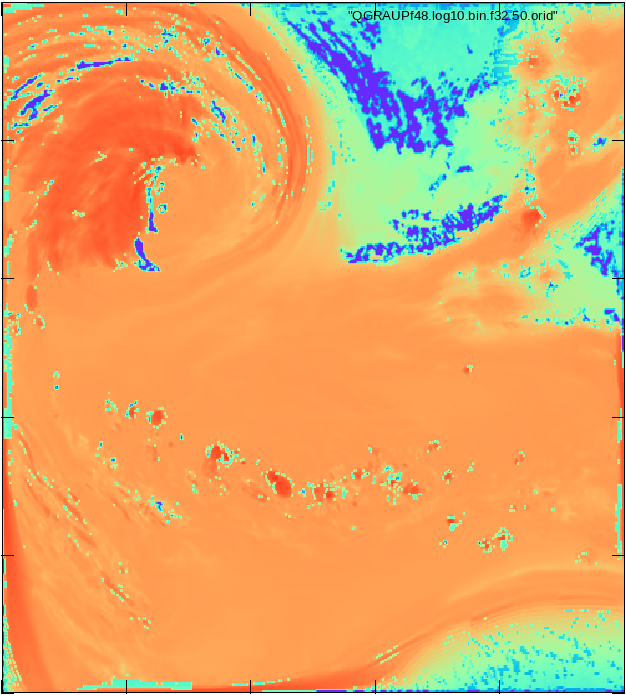}
    \caption{QGRAUP}
    \label{fig:b}
  \end{subfigure}
  \begin{subfigure}[t]{0.24\columnwidth}
    \centering
    \includegraphics[width=\linewidth]{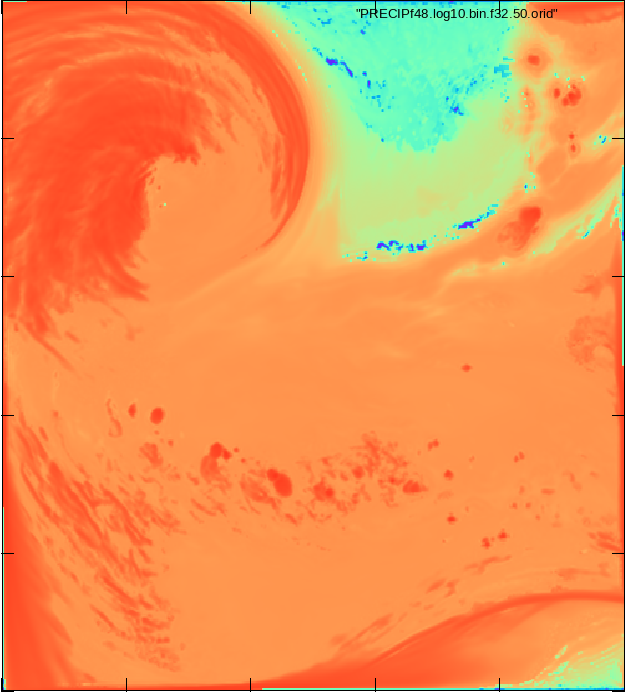}
    \caption{PRECIP}
    \label{fig:c}
  \end{subfigure}
  \vspace{-5pt}
  \caption{Similar pattern across four fields in the Hurricane application.}
  \label{fig:2}
  \vspace{-10pt}
\end{figure}

Motivated by this observation, we propose a \textbf{two-stage compression quality prediction framework} (Fig.~\ref{fig:3}) that decouples the computationally expensive \textit{feature extraction backbone} from the \textit{quality metric prediction head}. This design offers several key advantages over the approach adopted in prior work~\cite{mumenin2024qualitynet}:
\textbf{(1) Efficient training}\textemdash the feature extraction backbone is trained once across all data fields within an application, significantly reducing training and storage overhead;
\textbf{(2) Flexible design}\textemdash compression researchers can independently refine the architectures of the feature extractor or the prediction head to enhance performance without incurring the full-model retraining costs; and
\textbf{(3) Modular inference}\textemdash users can compose customized prediction frameworks by combining pre-trained components, like assembling LEGO pieces, to address diverse prediction tasks.
The remainder of this section describes the design of the feature extraction backbone and the metric prediction head, highlighting how their integration enables scalable and accurate compression quality prediction.

\subsection{Feature Extraction Backbone}\label{sec:3.3}
The feature extraction backbone, implemented as a Data Feature Extraction Network (DFE-NN), is designed to capture representative features from high-dimensional scientific data, forming the foundation for downstream compression quality prediction. Prior work by Mumenin et al.~\cite{mumenin2024qualitynet} demonstrated that CNN-based architectures are effective for extracting spatial features from 2D scientific datasets. Building on this insight, we explore whether more advanced model architectures can further enhance feature extraction for 3D data.
Recent studies indicate that CNN-based models, such as ResNet~\cite{8821313,lu2023multiscale,guo2024improved} and U-Net~\cite{ronneberger2015u}, remain the dominant choice for 3D volumetric scientific data due to their efficiency in learning local spatial patterns and their robustness under limited training data. Their convolutional inductive biases facilitate the capture of hierarchical spatial dependencies, enabling stable and effective feature extraction for metrics such as CR, PSNR, and SSIM even with relatively small datasets. In contrast, transformer-based architectures such as ViT~\cite{Srinadh} and ViViT~\cite{arnab2021vivit} excel at modeling long-range dependencies and global correlations through self-attention mechanisms, making them appealing for complex spatiotemporal data in domains like climate, astrophysics and MRI~\cite{forigua2022superformer}. However, these models typically require large-scale datasets and extensive computational resources to prevent overfitting, which limits their practicality in scientific workflows.

Given that our backbone is trained per scientific application rather than as a large-scale foundational model, assembling massive datasets (on the order of $\sim$10,000 samples) is infeasible. Therefore, in this work, we adopt a CNN-based ResNet-152 backbone, modified to operate on 3D data by replacing all 2D convolutional and pooling layers with their 3D counterparts. This extension enables the model to capture cross-dimensional correlations across width, height, and depth, ensuring efficient and reliable feature extraction for subsequent quality metric prediction.

\begin{figure}[t]
    \centering
    \includegraphics[width=.8\columnwidth]{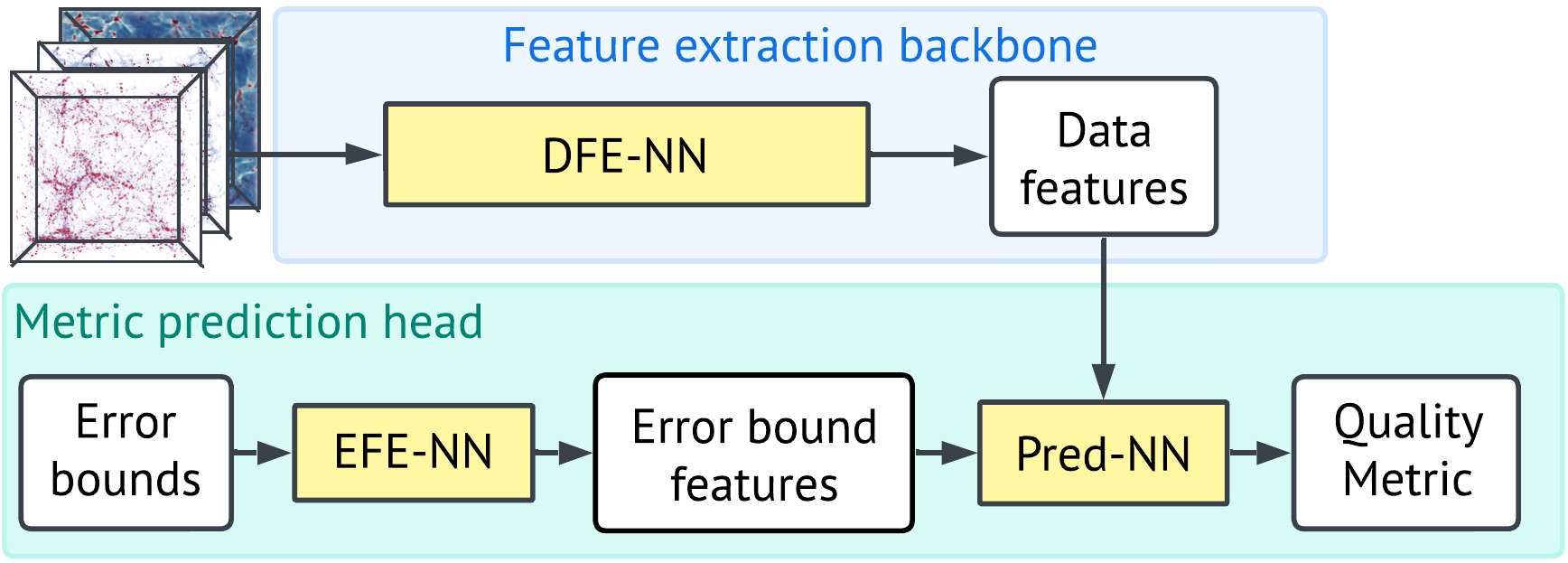}
    \caption{Two-stage compression quality prediction framework. Highlighted parts are neural network components of our surrogate model.}
    \vspace{-15pt}
    \label{fig:3}
\end{figure}

\subsection{Metric Prediction Head}\label{sec:3.4}
Compression quality is jointly influenced by the input data and the specified error bound, requiring the metric prediction head to integrate features from both sources. While the input data features are extracted by the DFE-NN, an important design question arises: how can a scalar error bound be effectively represented and fused with data features to enable accurate quality metric prediction? To address this, we design the metric prediction head with two key components\textemdash the Error-bound Feature Extraction Network (EFE-NN) and the Prediction Network (Pred-NN)\textemdash as illustrated in Fig.~\ref{fig:3}.

\noindent \underline{\textit{(1) Error-bound feature extraction:}}
In error-bounded lossy compression workflows, the error bound is a single scalar value, offering limited representational capacity for capturing its complex influence on data quality. To overcome this limitation, we employ a learnable embedding strategy~\cite{NEURIPS2022_9e9f0ffc} by introducing the EFE-NN, which transforms the scalar error bound into a high-dimensional embedding. The EFE-NN consists of two multi-layer perceptrons (MLPs) with 128 and 256 neurons, respectively. This lightweight yet expressive architecture effectively encodes the scalar error bound into semantically rich representations that enhance downstream metric prediction accuracy.

\noindent \underline{\textit{(2) Prediction feature combination:}}
Once the data and error-bound features are extracted, they must be effectively fused for prediction. While complex fusion techniques have been explored in computer vision~\cite{late_fuison}, many rely on intensive cross-feature interactions that risk distorting the learned data features—undesirable in our setting, where quality metrics are primarily governed by variations in the error bound. To preserve feature integrity, we adopt a streamlined, Siamese-inspired fusion strategy~\cite{NIPS1993_288cc0ff}. Both the data features and the embedded error-bound features are fed into the Pred-NN, which integrates them in a balanced manner to predict the desired quality metrics. This design promotes generalization across compressors and scientific applications, with varying relationships between error bounds and quality metrics.

\begin{figure}[t]
    \centering
    \begin{subfigure}[t]{\columnwidth}
        \centering
        \includegraphics[width=.9\columnwidth]{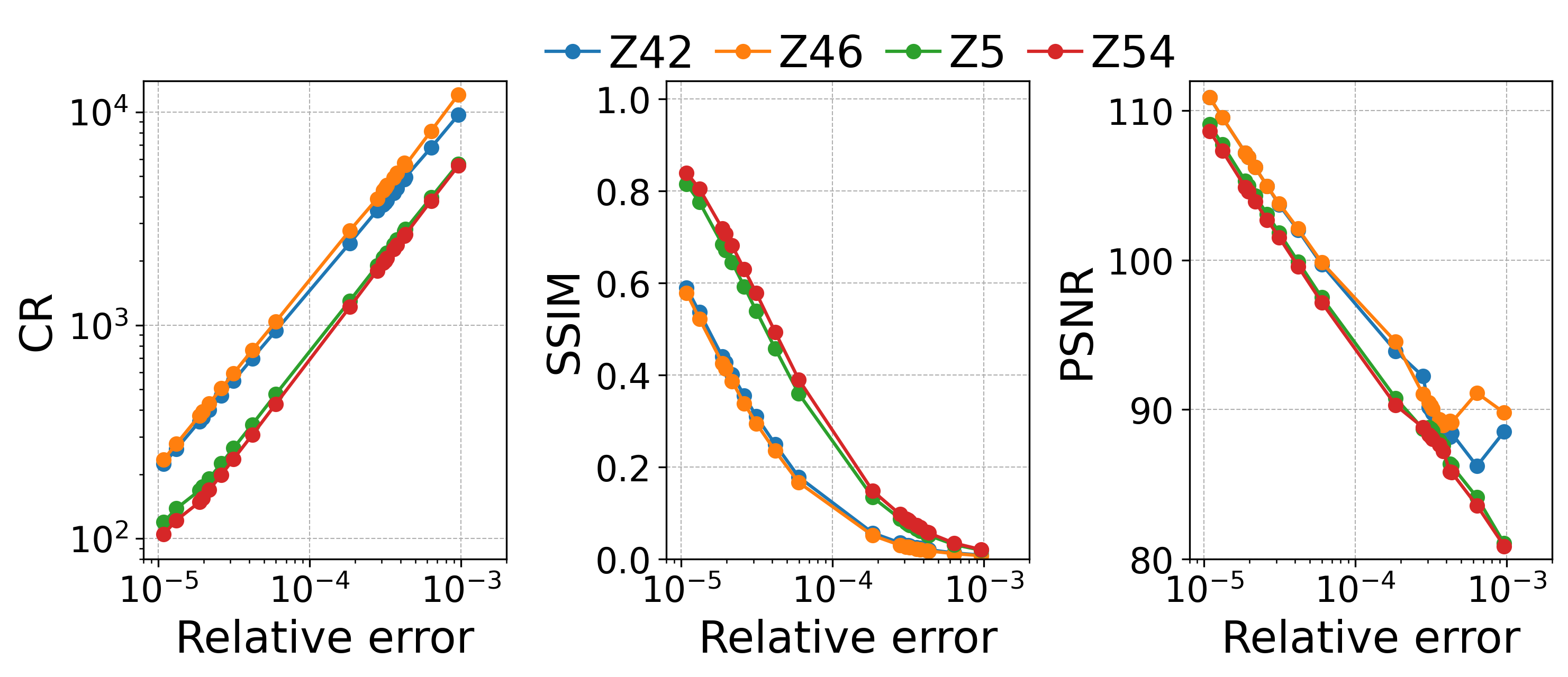}
        \vspace{-5pt}
        \caption{SZ3}
        \label{fig:4-bd}
    \end{subfigure}
    \hfill
    \begin{subfigure}[t]{\columnwidth}
        \centering
        \includegraphics[width=.9\columnwidth]{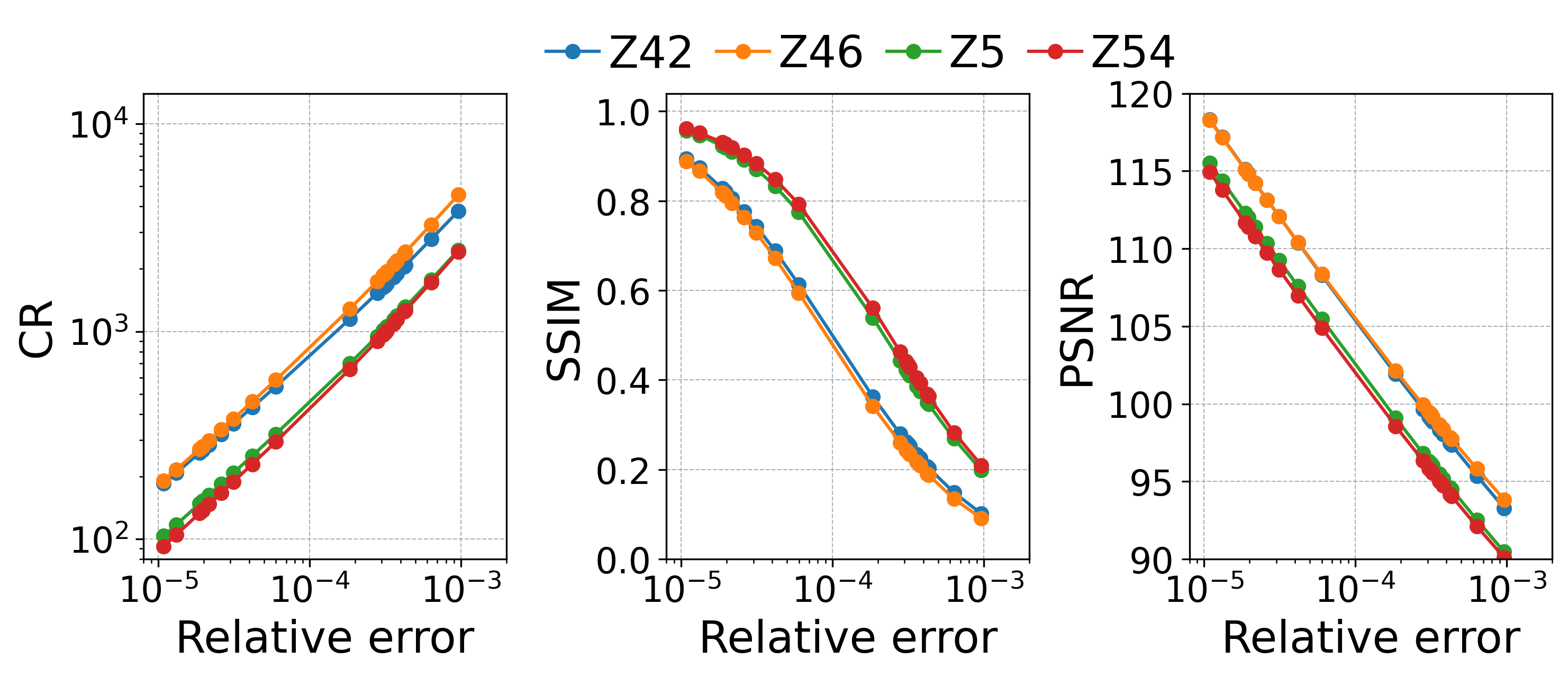}
        \vspace{-5pt}
        \caption{SPERR}
        \label{fig:4-dmd}
    \end{subfigure}
    \vspace{-6pt}
    \caption{Compression quality of Baryon density across four timesteps.}
    \vspace{-15pt}
    \label{fig:4}
\end{figure}

\subsection{Optimization: Generalization for Time-Evolving Data}\label{sec:3.5}
The previously discussed metric prediction head provides a simple yet effective means of estimating compression quality from combined data and error-bound features. However, in certain scenarios, particularly in the Nyx simulation, we observe significant degradation in prediction accuracy due to a noticeable distributional gap between training and testing data.

In the Nyx cosmology simulation, the timestep control is governed by the redshift parameter (\texttt{z}), with data snapshots recorded at specific redshifts. For example, timesteps \texttt{z42}, \texttt{z46}, \texttt{z5}, and \texttt{z54} correspond to redshifts 4.2, 4.6, 5.0, and 5.4, respectively. As the universe evolves, redshift decreases with increasing cosmic time, and Nyx performs hundreds to thousands of integration steps between redshift checkpoints. Consequently, data across timesteps exhibit substantial structural variation, leading to significant changes in compression behavior. As shown in Fig.~\ref{fig:4}, the CR, PSNR, and SSIM naturally cluster into two distinct groups (\texttt{z42}/\texttt{z46} and \texttt{z5}/\texttt{z54}).

This temporal variation poses a challenge for generalization when the prediction head is trained on \texttt{z42}, \texttt{z46}, and \texttt{z5}, but evaluated on \texttt{z54}. In this scenario, the training and testing data distributions differ significantly, causing predictive bias. This behavior reflects a broader challenge in many time-evolving scientific applications, where models must be trained on early timesteps and deployed on later ones that differ substantially in physical characteristics.

To mitigate this bias and enhance robustness, we adopt a \textbf{Mixture-of-Experts (MoE)} architecture~\cite{Noammoe} on top of the existing Pred-NN, as illustrated in Fig.~\ref{fig:5}. Unlike a single unified network, the MoE design distributes the prediction task across multiple expert modules, each specializing in distinct subregions of the input space. Then, a \textit{router module} dynamically aggregates the expert outputs according to learned weights, employing soft gating to ensure stable combination of predictions and avoid the instability associated with sparse expert selection~\cite{puigcerver2024sparsesoftmixturesexperts,Cai_2025}. In this work, we use four experts, balancing expressiveness with computational efficiency.

\begin{figure}[t]
    \centering
    \includegraphics[width=.9\columnwidth]{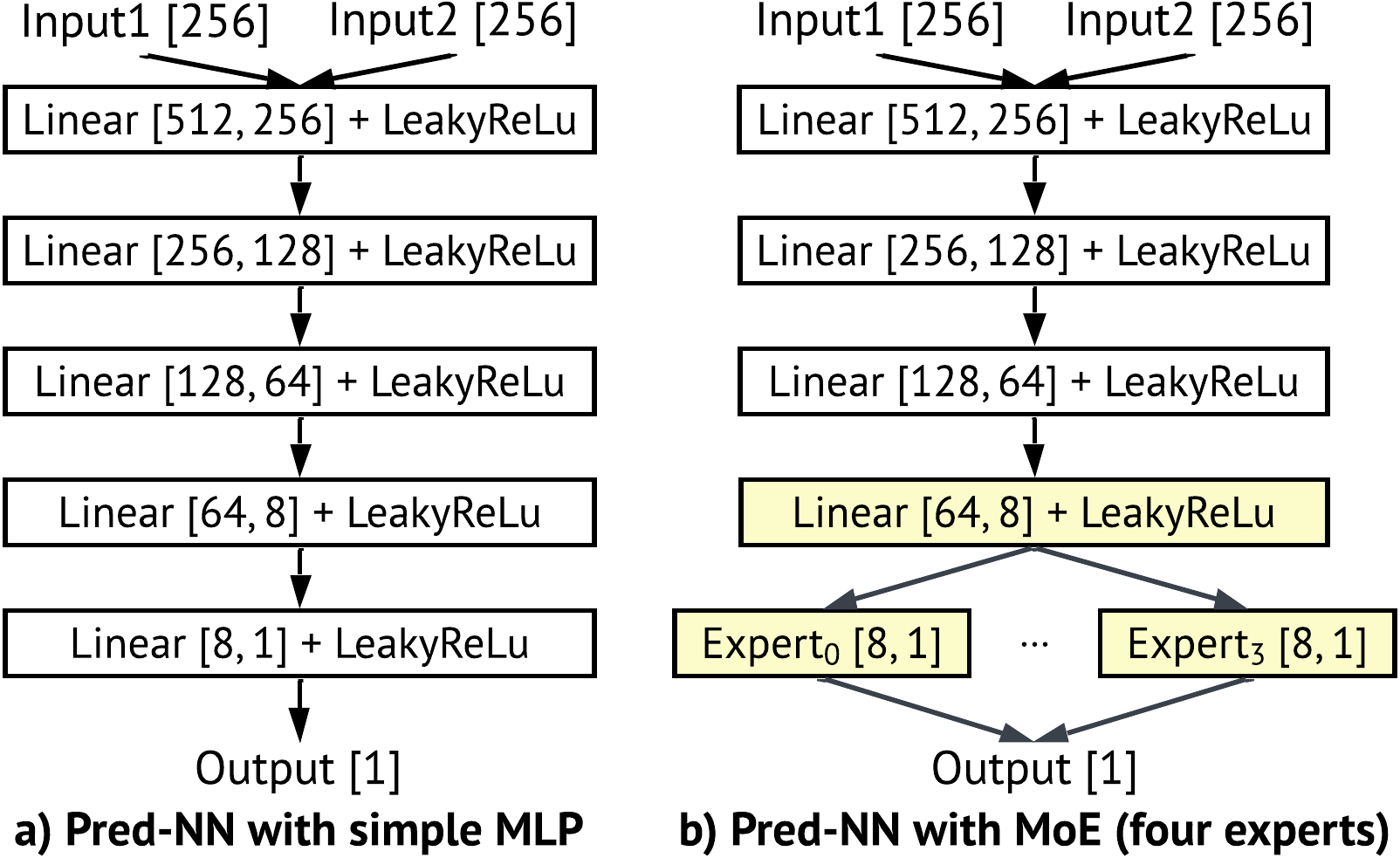}
    \caption{The architecture of Pred-NN. Highlighted components in b) represent the MoE design.}
    \vspace{-12pt}
    \label{fig:5}
\end{figure}

The design choices in metric prediction head enables robust prediction across timesteps, compressors, and scientific applications. Despite the modular composition (EFE-NN, Pred-NN, and MoE), the overall computational footprint remains lightweight. Furthermore, all components are trained jointly in an end-to-end manner, ensuring stable optimization and efficient convergence.

\section{Evaluations} \label{sec:eval}

In this section, we evaluate the performance of our surrogate-based compression quality prediction framework on real-world scientific datasets from the Scientific Data Reduction Benchmark~\cite{sdrbench}, including Nyx cosmology simulation~\cite{Almgren_2013}, Hurricane Isabel climate simulation~\cite{hurricane-isabel}, Miranda turbulence simulations~\cite{miranda}, and the RTM seismic dataset~\cite{rtm}. For the Miranda application, the full-scale dataset is partitioned into subdomains of size $256\times384\times384$ to create disjoint training and test samples. Table~\ref{tab:data} summarizes the details of these datasets.

\begin{table}[h]
\setlength{\tabcolsep}{3pt}
\renewcommand{\arraystretch}{1.2}
\centering
\caption{Datasets used in this work.}
\vspace{-5pt}
\label{tab:data}
\resizebox{\columnwidth}{!}{%
\begin{tabular}{cccccc}
\hline \hline
\textbf{Application} & \textbf{\# Timesteps} & \textbf{\# Fields} & \textbf{Dimension}      & \textbf{Size} \\ \hline
Nyx        & 4    & 4   & $512\times512\times512$ &  2.04 GB \\ \hline
Hurricane  & 48   & 13  & $100\times500\times500$ & 58.50 GB \\ \hline
Miranda    & 1    & 1   & $3072\times3072\times3072$ & 109.00 GB \\ \hline
RTM        & 250  & 1   & $235\times449\times449$ & 45.00 GB \\ \hline \hline

\end{tabular}
}
\end{table}

Our evaluation proceeds in several stages. We first assess the prediction accuracy of the proposed framework on individual data fields (Section~\ref{sec:4.2}) and compare the results against existing methods (Section~\ref{sec:4.3}). We then perform ablation studies (Section~\ref{sec:4.5}) to evaluate the impact of key design choices, including the training overhead reduction using the two-stage framework and the MoE module for handling time-evolving data. We also demonstrate the generalization capability of the proposed approach across multiple data fields within each application, on the Nyx and Hurricane datasets.

\subsection{Experimental Setup} \label{sec:4.1}

\subsubsection{Hardware and Software Setup}
All experiments, including compression, decompression, quality metric computation, and surrogate model training, are conducted on a local cluster equipped with 48 Intel(R) Xeon(R) Silver 4214 CPUs and 8 NVIDIA Quadro RTX 5000 GPUs.

Table~\ref{tab:compression-expr} summarizes the software used in our experiments. We evaluate five widely adopted lossy compressors---SZ2~\cite{di2016fast,tao2017significantly,liang2018error,zhao_significantly_2020}, SZ3~\cite{SZ3,Zhao2021,liu2024high}, SZx~\cite{SZx}, ZFP~\cite{Lindstrom2014}, and SPERR~\cite{sperr}---which collectively cover diverse compression philosophy and offer varying trade-offs between compression ratio and reconstruction quality. All compression tasks are executed via the LibPressio framework~\cite{libpressio}, which provides a unified interface and standardized performance metrics. Quality metrics are computed using the Quick Compression Analysis Toolkit (QCAT)~\cite{szcompressor_qcat} to ensure consistent and reproducible evaluation across compressors and datasets.

\begin{table}[t]
\setlength{\tabcolsep}{4pt}
\centering
\caption{Software used in compression experiments.}
\label{tab:compression-expr}
\vspace{-5pt}
\resizebox{\columnwidth}{!}{%
\begin{tabular}{lll}
\hline \hline
\multicolumn{1}{c}{\textbf{Name}} & \multicolumn{1}{c}{\textbf{Version}} & \multicolumn{1}{c}{\textbf{URL}} \\ \hline
SZ2        & v2.1.12 & \texttt{https://github.com/szcompressor/SZ } \\ \hline
SZ3        & v3.2.1  & \texttt{https://github.com/szcompressor/SZ3}     \\ \hline
SZx        & v1.1.0  & \texttt{https://github.com/szcompressor/SZx}     \\ \hline
ZFP        & v1.0.1  & \texttt{https://github.com/LLNL/zfp}             \\ \hline
SPERR      & v0.8.1  & \texttt{https://github.com/NCAR/SPERR}           \\ \hline
LibPressio & v1.0.2  & \texttt{https://github.com/robertu94/libpressio }\\ \hline
QCAT       & v1.7.1  & \texttt{https://github.com/szcompressor/qcat }   \\ \hline \hline
\end{tabular}%
}
\vspace{-15pt}
\end{table}

\subsubsection{Compression Configurations} \label{sec:4.1.2}
Compression quality is highly sensitive to the configurations, particularly the choice of error bounds. In this study, we configure error bounds to balance between \textit{{achieving meaningful compression ratios}} and \textit{preserving sufficient fidelity for downstream scientific analysis}. Guided by domain experts, we select relative error bounds within the following ranges: Nyx [1e-5, 1e-3], Hurricane [1e-5, 1e-2], Miranda [1e-4, 1e-2], and RTM [1e-4, 1e-3].

\subsubsection{Model Training and Testing Configurations}
We employ a two-stage training procedure, which aligns with our modular design. In the first stage, we train a shared feature extraction backbone across all fields of each application. In the second stage, the backbone is frozen, and task-specific prediction heads are trained independently.

\noindent \underline{\textit{Phase 1: Backbone Training}}  
To capture shared patterns across data fields, the backbone must be trained on a representative set of samples. Three challenges arise in this process:  
{\textbf{(i) Limited training diversity}}---acquiring diverse timesteps is often in feasible~\cite{DBLP:journals/corr/ZhuVFR15}.  
{\textbf{(ii) GPU memory constraints}}---the large scale of scientific data complicates mini-batch data loading~\cite{KANDEL2020312}.  
{\textbf{(iii) Varying input dimensions}}---varying simulation output dimensions (e.g., Nyx data ranging from $512^3$ to $2048^3$~\cite{friesen_situ_2016}) hinders direct ingestion for CNN-based models.

To address these challenges, we randomly sample small 3D blocks from full-resolution data. This downsampling strategy reduces and unifies input dimensionality, increases the number of training samples, and retains local spatial structures without introducing partitioning bias. The sampled blocks are normalized using min–max scaling. The backbone is trained for 250 epochs with the Adam optimizer, an initial learning rate of 0.01 with scheduler decay, and MSE loss.

\noindent \underline{\textit{Phase 2: Prediction Head Training}}  
Error bounds span several orders of magnitude (e.g., $ 10^{-5}$ to $10^{-1}$), which can impede model convergence as model gives disproportionate attention to the large errors, effectively ignoring the much smaller ones. To mitigate this issue, we apply a logarithmic normalization to error bounds before feeding into EFE-NN. Each prediction head is trained for 150 epochs using the Adam optimizer, a learning rate scheduler initialized at 0.01, and RMSE loss.

To ensure fair evaluation, all datasets are divided into disjoint training and testing subsets. For the Nyx dataset, the first three timesteps (\texttt{z42}, \texttt{z46}, \texttt{z5}) are used for training and the last one (\texttt{z54}) for testing to evaluate temporal generalization. For the other applications, timesteps are split using an \textit{odd–even partitioning strategy}, where odd-indexed timesteps are used for training and even-indexed ones for testing.

We use the Percentage Error (PE) to quantify the relative prediction error against ground truth:
\begin{equation*}
    \text{PE} = \frac{\text{Orig} - \text{Pred}}{\text{Orig}} \times 100
\end{equation*}
and Mean Absolute Percentage Error (MAPE) to measure the average percentage error across all error bounds:
\begin{equation*}
    \text{MAPE}=\frac{1}{n} \sum \left|\frac{\text{Orig}-\text{Pred}}{\text{Orig}}\right| \times 100
\end{equation*}
where $n$ is the number of evaluated error bounds.

\begin{table*}[h!]
\setlength{\tabcolsep}{3pt}
\renewcommand{\arraystretch}{1.2}
\caption{Compression quality prediction error of DeepCQ (in \%).}
\label{tab:qnet-pred-err}
\vspace{-5pt}
\begin{threeparttable}
\resizebox{\textwidth}{!}{%
\begin{tabular}{cccccccccccccccccccccc}
\hline \hline
\multirow{2}{*}{\textbf{Application}} 
& \multirow{2}{*}{\textbf{Fields}} &  
& \multicolumn{3}{c}{\textbf{SZ2}}  &  
& \multicolumn{3}{c}{\textbf{SZ3}}   &  
& \multicolumn{3}{c}{\textbf{SZx}} &  
& \multicolumn{3}{c}{\textbf{ZFP}} &  
& \multicolumn{3}{c}{\textbf{SPERR}}  \\ \cline{4-6} \cline{8-10} \cline{12-14} \cline{16-18} \cline{20-22} & &  
& CR & PSNR & SSIM &  
& CR & PSNR & SSIM &  
& CR & PSNR & SSIM &  
& CR & PSNR & SSIM &  
& CR & PSNR & SSIM \\ \hline \hline
\multirow{4}{*}{Nyx}  
& Baryon density      &  & 5.33      & 0.74 & 6.21         &  & 5.82       & 1.17      & 9.05        &  & 9.64 & 1.74 & 6.40     &  & 9.45      & 1.08         & 6.52      &  & 4.25      & 3.02 & 5.92   \\
& Dark matter density &  & 4.96      & 1.44 & 4.45         &  & 3.54       & 1.18      & 2.84        &  & 3.10 & 1.70 & 3.29     &  & 5.07      & 1.60         & 1.86      &  & 3.69      & 2.33 & 6.23   \\
& Temperature         &  & 3.54      & 3.61 & 0.94         &  & 3.13       & 3.71      & 1.10        &  & 2.34 & 2.84 & 1.77     &  & 4.51      & 3.30         & 1.63      &  & 1.51      & 3.72 & 0.67   \\
& Velocity x          &  & 1.38      & 1.59 & 2.13         &  & 2.20       & 1.66      & 2.18        &  & 1.89 & 1.97 & 2.10     &  & 2.84      & 1.57         & 2.00      &  & 4.04      & 2.32 & 2.04   \\ \hline
\multirow{13}{*}{Hurricane}    
& CLOUD               &  & 2.19      & 1.96 & 1.24         &  & 1.57      & 1.00      & 0.39         &  & 2.74 & 3.08 & 0.40     &  & 3.35      & 2.87         & 0.22      &  & 3.95      & 1.75 & 0.67   \\
& QCLOUD              &  & 3.57      & 1.53 & 1.45         &  & 3.46      & 0.83      & 1.72         &  & 6.30 & 2.64 & 0.86     &  & 4.59      & 1.67         & 1.06      &  & 3.81      & 1.80 & 1.16   \\
& PRECIP              &  & 1.93      & 1.88 & 2.30         &  & 9.57      & 1.52      & 2.16         &  & 6.74 & 3.66 & 1.31     &  & 6.29      & 1.97         & 2.30      &  & 2.29      & 3.18 & 2.83   \\
& QGRAUP              &  & 3.97      & 2.20 & 11.39        &  & 8.78      & 0.82      & 18.90        &  & 3.26 & 3.32 & 7.70     &  & 5.04      & 1.43         & 13.21     &  & 4.13      & 1.61 & 17.19  \\
& QICE                &  & 7.17      & 2.53 & 0.35         &  & 9.76      & 1.39      & 0.23         &  & 1.82 & 2.68 & 0.23     &  & 5.19      & 2.66         & 0.07      &  & 10.80     & 2.59 & 0.70   \\
& QRAIN               &  & 4.78      & 2.79 & 10.73        &  & 3.38      & 1.72      & 6.75         &  & 3.96 & 3.82 & 2.43     &  & 4.68      & 2.52         & 8.24      &  & 6.15      & 2.64 & 11.15  \\
& QSNOW               &  & 8.88      & 1.43 & 5.22         &  & 2.53      & 1.45      & 1.31         &  & 3.80 & 4.20 & 2.95     &  & 6.12      & 2.76         & 3.27      &  & 8.87      & 1.71 & 4.66   \\
& QVAPOR              &  & 6.07      & 1.91 & 0.90         &  & 10.78     & 1.46      & 0.64         &  & 3.20 & 5.01 & 0.89     &  & 4.38      & 2.14         & 0.53      &  & 4.86      & 2.60 & 0.38   \\
& P                   &  & 2.11      & 1.98 & 0.34         &  & 5.24      & 1.53      & 0.24         &  & 5.80 & 3.68 & 0.25     &  & 5.02      & 2.35         & 0.52      &  & 3.78      & 2.43 & 0.11   \\
& TC                  &  & 5.91      & 2.97 & 1.49         &  & 11.39     & 1.31      & 0.05         &  & 9.25 & 4.41 & 0.91     &  & 4.32      & 2.12         & 0.16      &  & 7.76      & 0.77 & 0.02   \\
& U                   &  & 3.10      & 2.07 & 0.55         &  & 3.61      & 0.85      & 0.12         &  & 4.84 & 3.57 & 0.12     &  & 4.36      & 2.14         & 0.07      &  & 3.17      & 1.06 & 0.53   \\
& V                   &  & 4.22      & 1.91 & 0.55         &  & 2.46      & 1.13      & 0.14         &  & 3.55 & 3.79 & 1.39     &  & 4.31      & 1.19         & 0.05      &  & 5.41      & 1.57 & 0.03   \\ 
& W                   &  & 11.40     & 2.72 & 0.87         &  & 9.34      & 1.03      & 0.56         &  & 7.63 & 2.34 & 2.60     &  & 4.25      & 3.22         & 0.25      &  & 11.25     & 2.64 & 0.72   \\ \hline
Miranda                   
& Density             &  & 11.05     & 1.54 & 0.89        &  & 9.81       & 0.64     & 0.19          &  & 3.52 & 1.66 & 0.39     &  & 12.75     & 1.88         & 0.50      &  & 4.09      & 0.55 & 0.17   \\ \hline
RTM                       
& Wavefield           &  & 2.24      & 1.24 & 1.59         &  & 3.96       & 2.08      & 1.66        &  & 2.98 & 1.51 & 1.38     &  & 3.37      & 1.49         & 0.96      &  & 3.67      & 2.23 & 3.59   \\ \hline \hline
\end{tabular}%
}
\begin{tablenotes}
    \footnotesize
    \item All prediction error values are averaged across all testing error bounds selected from the preferred range.
\end{tablenotes}
\end{threeparttable}
\vspace{-5pt}
\end{table*}

\subsection{Prediction Accuracy on Individual Fields}\label{sec:4.2}
Table~\ref{tab:qnet-pred-err} presents the comprehensive evaluation results, measured in MAPE, across all combinations of compressors, quality metrics, and data fields. Our framework demonstrates {\textbf{consistently high prediction accuracy}}, achieving a prediction error below 10\% in {\textbf{272 out of 285 test cases}}. This level of accuracy highlights the strong capability of the surrogate-based approach to learn complex interactions among input data, error bounds, and compressor behaviors. Furthermore, the modular design effectively isolates prediction tasks across compressors and quality metrics, avoiding cross-interference and enhancing scalability for real-world deployment.

\begin{mdframed}[linecolor=black, backgroundcolor=gray!2, roundcorner=6pt]
\textbf{\textit{Takeaway \#1:}} The proposed surrogate-based framework achieves exceptional prediction accuracy across diverse compressors, quality metrics, and scientific datasets, with a typical prediction error below 10\%.
\end{mdframed}

Among the limited cases where the prediction error exceeds 10\%, we identify three representative scenarios that contribute to reduced prediction accuracy, as illustrated in Fig.~\ref{fig:6}.
First, SSIM predictions on the QGRAUP field across compressors exhibit reduced accuracy, primarily due to variations in SSIM sensitivity to error bounds across timesteps. Results from compression experiments show that as timesteps progress, the SSIM range narrows, indicating gradual structural drift relative to the error bounds. Despite using an odd–even data partitioning strategy, the limited data volume restricts the model’s ability to generalize across temporal variations. 

Second, metrics predicted for ZFP display deviations due to the compressor’s non-differentiable behavior. Specifically, ZFP’s bitplane truncation of transform coefficients introduces discontinuities in the relationship between CR and error bounds. This block-wise truncation results in a stair-step CR curve that is inherently challenging for a continuous surrogate model to approximate, leading to mild over-smoothing in predictions (e.g., CR on the Density field).

Finally, CR prediction for SZ2 on the W field is affected by slight non-monotonic behavior, where the CR decreases around an error bound of approximately 2e–3. Such irregularities violate the monotonic assumption implicitly learned by the surrogate model that CR increases with relaxed error bounds, thus leading to minor prediction deviations.

\begin{figure}[t]
    \centering
    \begin{subfigure}{.33\columnwidth}
        \centering
        \includegraphics[width=\columnwidth]{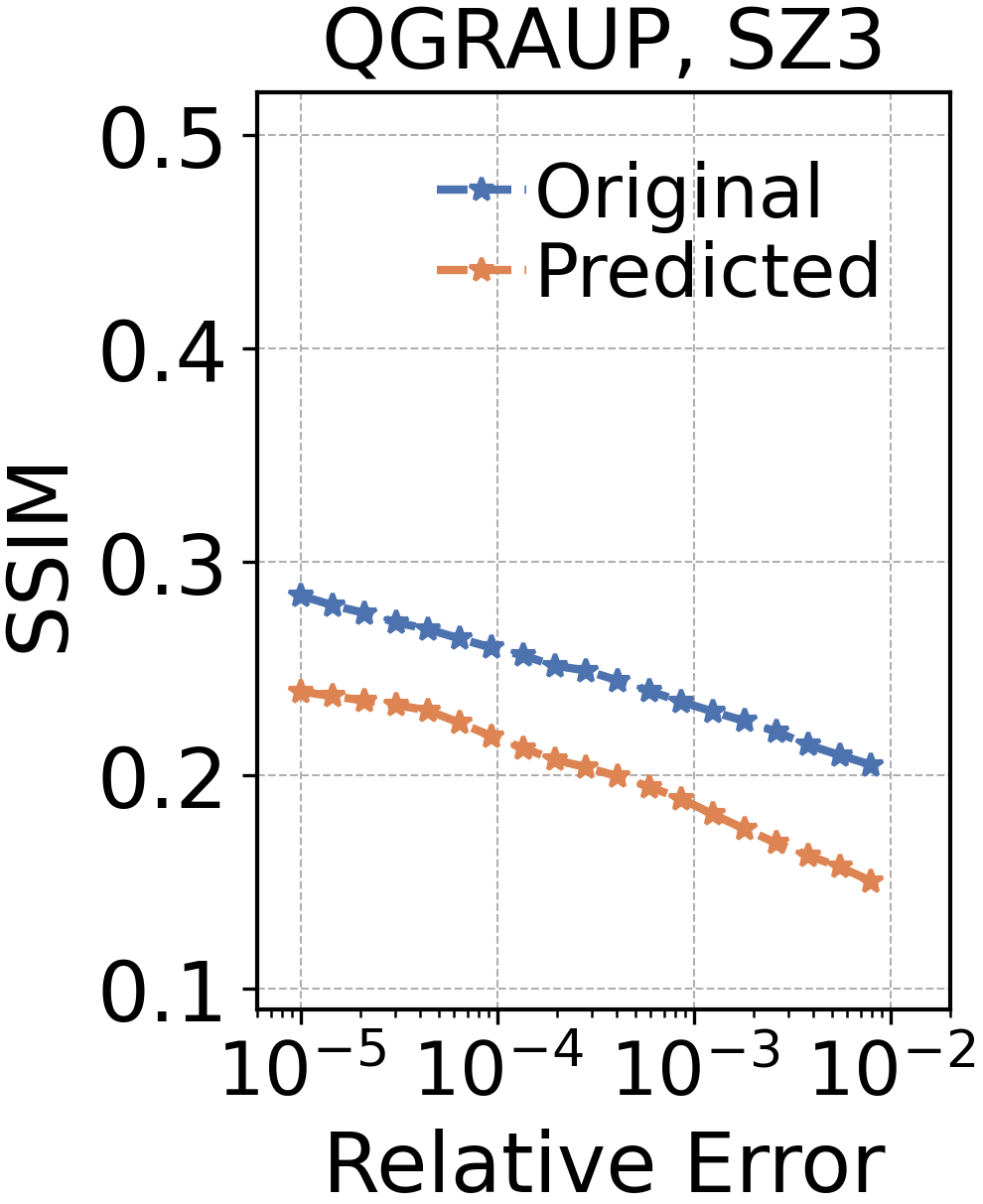}
        \vspace{-14pt}
    \end{subfigure}
    \begin{subfigure}{.315\columnwidth}
        \centering
        \includegraphics[width=\columnwidth]{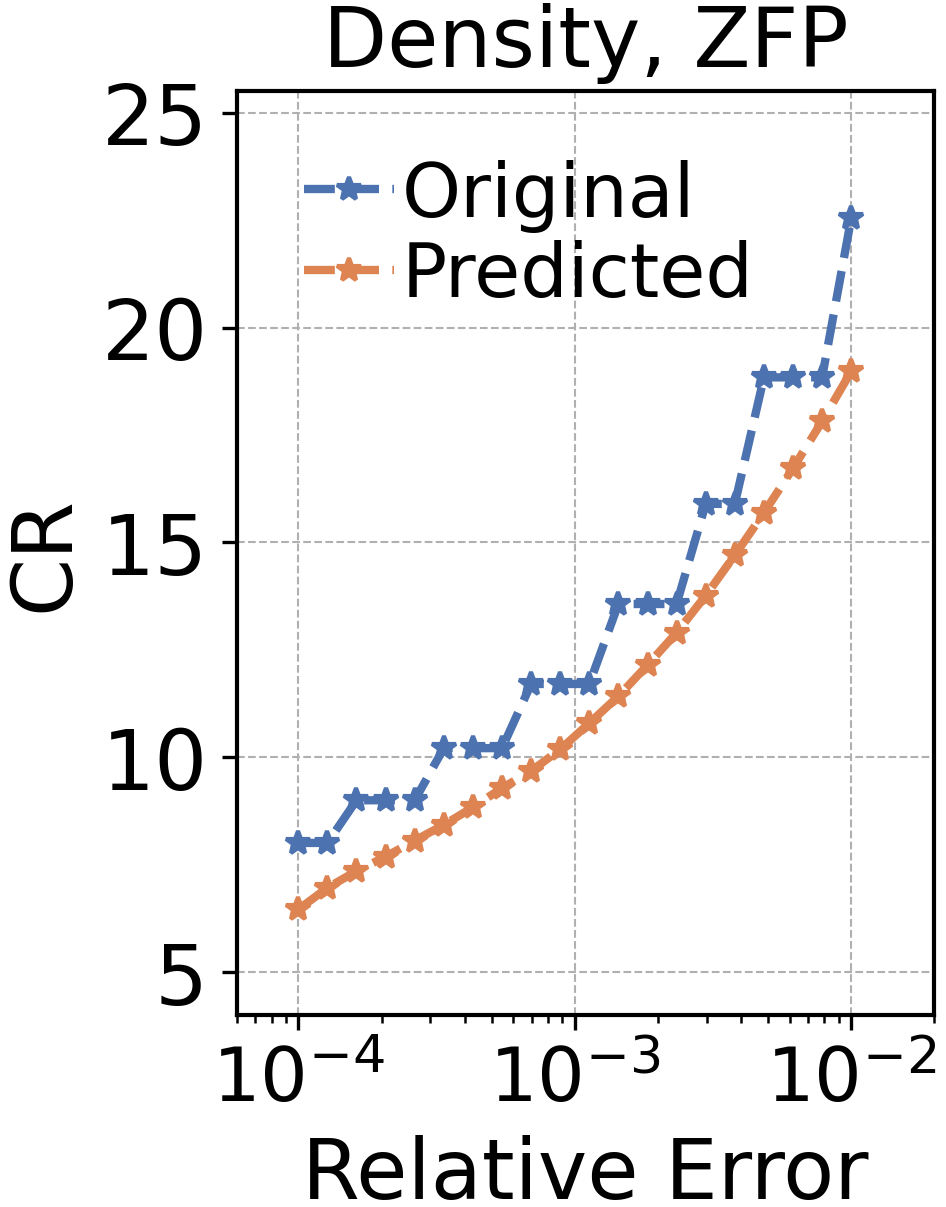}
        \vspace{-14pt}
    \end{subfigure}
    \begin{subfigure}{.32\columnwidth}
        \centering
        \includegraphics[width=\columnwidth]{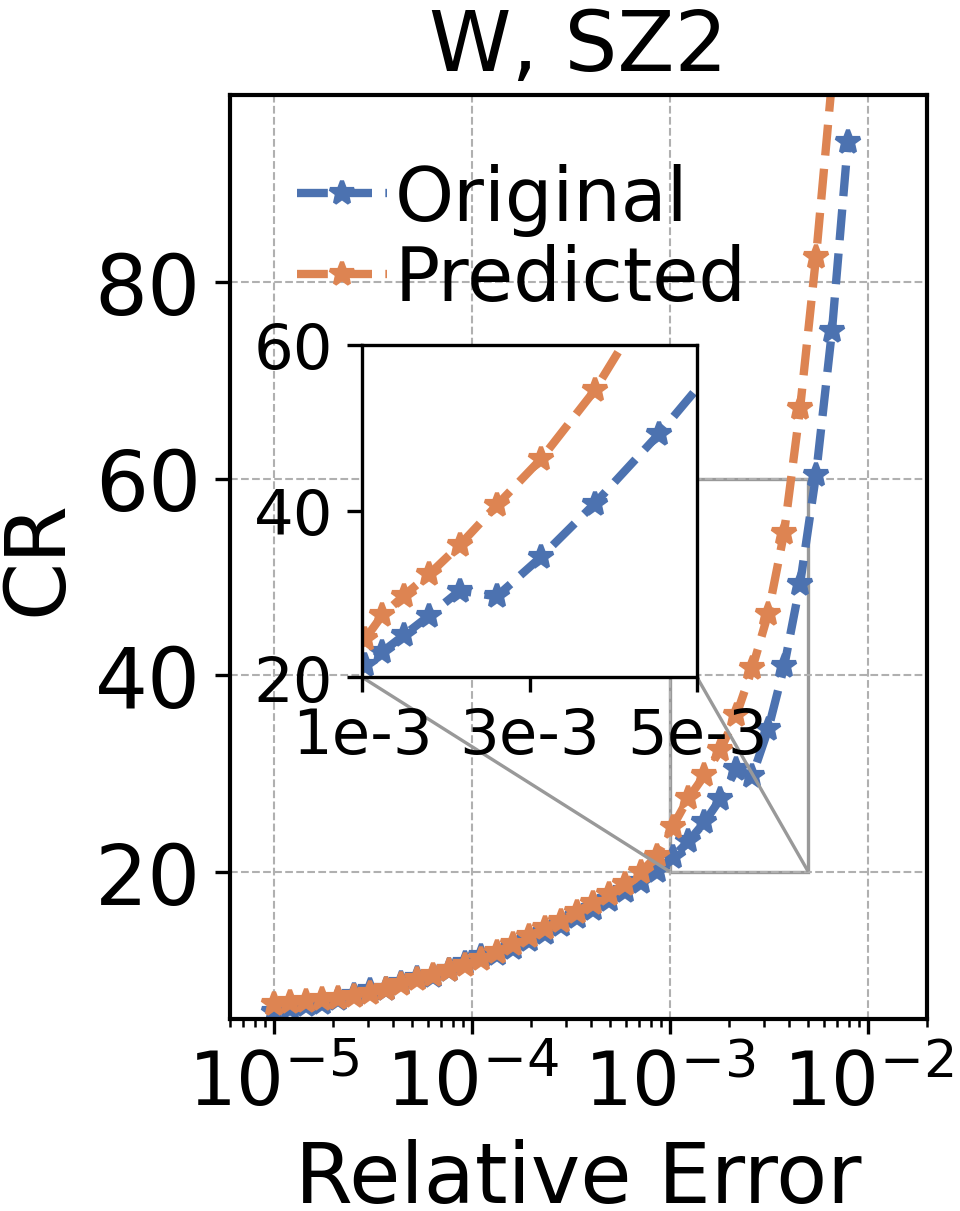}
        \vspace{-14pt}
    \end{subfigure}
    \vspace{-10pt}
    \caption{Three examples where prediction is less accurate. They represent different kinds of challenges for compression quality prediction.}
    \vspace{-15pt}
    \label{fig:6}
\end{figure}

\begin{mdframed}[linecolor=black, backgroundcolor=gray!2, roundcorner=6pt]
\textbf{\textit{Takeaway \#2}:} The intrinsic interaction among input data, lossy compressors, and error bounds complicates surrogate-based compression quality prediction. Scenarios such as \textit{quality variation between training and testing data} and \textit{non-monotonic and stair-step behavior in compression quality} highlight the need for customized modeling strategies to ensure robust predictive performance.
\end{mdframed}

\subsection{Comparison with Existing Methods} \label{sec:4.3}

\begin{table*}[h!]
\caption{CR prediction error of DeepCQ compared with existing approaches (in $\%$).}
\setlength{\tabcolsep}{3pt}
\label{tab:pred_comparison}
\renewcommand{\arraystretch}{1.2}
\vspace{-5pt}
\begin{threeparttable}
\resizebox{\textwidth}{!}{%
\begin{tabular}{cccccccccccc}
\hline \hline
\multirow{2}{*}{\textbf{Fields}} 
& \multicolumn{2}{c}{\textbf{SZ2}} &  
& \multicolumn{4}{c}{\textbf{SZ3}} &  
& \multicolumn{3}{c}{\textbf{ZFP}} \\ \cline{2-3} \cline{5-8} \cline{10-12} 
& Tao2019~\cite{tao2019optimizing}& DeepCQ &  
& Khan2023~\cite{khan_secre_2023} & Tao2019~\cite{tao2019optimizing} & Jin2022~\cite{jin2022improving} & DeepCQ &
& Khan2023~\cite{khan_secre_2023} & Tao2019~\cite{tao2019optimizing} & DeepCQ \\ \hline \hline
Baryon density            & 99.26  & \underline{5.86} &     & 91.83 & 97.94  & 73.58 & \underline{6.80}  &     & \underline{0.06} & 7.02    & 9.53              \\
Dark matter density       & 94.35  & \underline{4.99} &     & 38.69 & 82.77  & 35.86 & \underline{3.70}  &     & \underline{0.02} & 4.04    & 5.03              \\
Temperature               & 99.26  & \underline{3.59} &     & 29.65 & 81.77  & 28.95 & \underline{3.28}  &     & \underline{0.02} & 2.30    & 4.46              \\
Velocity x                & 99.26  & \underline{1.70} &     & 12.61 & 76.41  & 13.88 & \underline{2.28}  &     & \underline{0.02} & 2.38    & 2.66              \\
CLOUD                     & 77.32  & \underline{2.46} &     & 62.77 & 90.79  & 36.37 & \underline{1.73}  &     & 78.40            & 29.13   & \underline{3.11}  \\
QCLOUD                    & 78.34  & \underline{3.59} &     & 76.36 & 93.32  & 36.12 & \underline{3.54}  &     & 79.08            & 149.24  & \underline{4.52}  \\
PRECIP                    & 81.17  & \underline{2.20} &     & 82.77 & 96.05  & 46.42 & \underline{9.45}  &     & 83.64            & 20.12   & \underline{6.38}  \\
QGRAUP                    & 85.12  & \underline{3.57} &     & 89.66 & 97.43  & 49.89 & \underline{6.29}  &     & 87.02            & 23.91   & \underline{5.53}  \\
QICE                      & 56.41  & \underline{7.57} &     & 63.32 & 89.54  & 32.91 & \underline{9.74}  &     & 76.86            & 314.27  & \underline{4.57}  \\
QRAIN                     & 72.07  & \underline{4.57} &     & 85.91 & 95.46  & 42.54 & \underline{3.55}  &     & 81.95            & 104.37  & \underline{4.54}  \\
QSNOW                     & 65.09  & \underline{9.10} &     & 74.67 & 93.48  & 43.95 & \underline{2.29}  &     & 78.67            & 40.79   & \underline{6.40}  \\
QVAPOR                    & 66.01  & \underline{5.95} &     & 56.98 & 86.89  & 24.81 & \underline{10.85} &     & 56.76            & 43.57   & \underline{5.07}  \\
P                         & 90.17  & \underline{2.15} &     & 75.90 & 95.45  & 47.48 & \underline{5.31}  &     & 62.89            & 61.30   & \underline{4.51}  \\
TC                        & 86.27  & \underline{6.35} &     & 62.54 & 92.51  & 32.45 & \underline{11.17} &     & 54.58            & 52.51   & \underline{4.33}  \\
U                         & 87.54  & \underline{2.91} &     & 59.92 & 92.26  & 26.26 & \underline{3.73}  &     & 55.89            & 55.97   & \underline{4.45}  \\
V                         & 87.76  & \underline{3.99} &     & 60.69 & 92.32  & 26.45 & \underline{2.39}  &     & 55.71            & 56.26   & \underline{4.41}  \\
W                         & 85.06  & \underline{11.20}&     & 57.30 & 91.15  & 27.14 & \underline{9.26}  &     & 44.77            & 46.03   & \underline{4.64}  \\
Density                   & 86.73  & \underline{11.05}&     & 84.98 & 95.61  & 51.96 & \underline{9.81}  &     & 77.96            & 79.32   & \underline{12.74} \\
RTM                       & 36.78  & \underline{2.24} &     & 79.76 & 95.31  & 92.87 & \underline{3.96}  &     & 79.44            & 125.24  & \underline{3.37}  \\ \hline \hline
\end{tabular}%
}
\begin{tablenotes}
    \footnotesize
    \item All prediction error values are averaged across 20 error bounds selected from the preferred range. 
    \item The best-performing method in each test case is underlined.
\end{tablenotes}
\end{threeparttable}
\vspace{-10pt}
\end{table*}


We compare the prediction error and time cost of DeepCQ against existing approaches. Since prior approaches primarily focus on CR prediction, our comparison is limited to this metric. Table~\ref{tab:pred_comparison} reports the MAPE of DeepCQ with Tao2019~\cite{tao2019optimizing}, Khan2023~\cite{khan_secre_2023}, and Jin2022~\cite{jin2022improving} for SZ2, SZ3, and ZFP.

Results demonstrate that \textbf{{DeepCQ achieves the lowest prediction error in 53 out of 57 test cases}}, as highlighted in Table~\ref{tab:pred_comparison}, with the only exception with ZFP, where Khan2023 and Tao2019 slightly outperform DeepCQ across the four Nyx data fields. This is attributed to ZFP’s block-wise compression mechanism, which produces highly regular patterns that sampling-based approaches can capture effectively when sampling is performed in a consistent block-wise manner.

It is worth noting that Tao2019 was primarily designed for online compressor selection between SZ2 and ZFP and therefore focuses on order-of-magnitude estimation rather than fine-grained accuracy. Moreover, due to algorithmic differences between SZ2 and SZ3, Tao2019 naturally exhibits reduced accuracy when applied to SZ3.

In addition, Khan2023 demonstrates variable prediction accuracy across datasets. Specifically, its performance degrades significantly as the sampling ratio decreases from 50\% to 10\%, particularly for the Hurricane dataset, where distinct data values across locations undermines the effectiveness of uniform sampling.

\begin{figure}[t]
    \centering
    \includegraphics[width=\columnwidth]{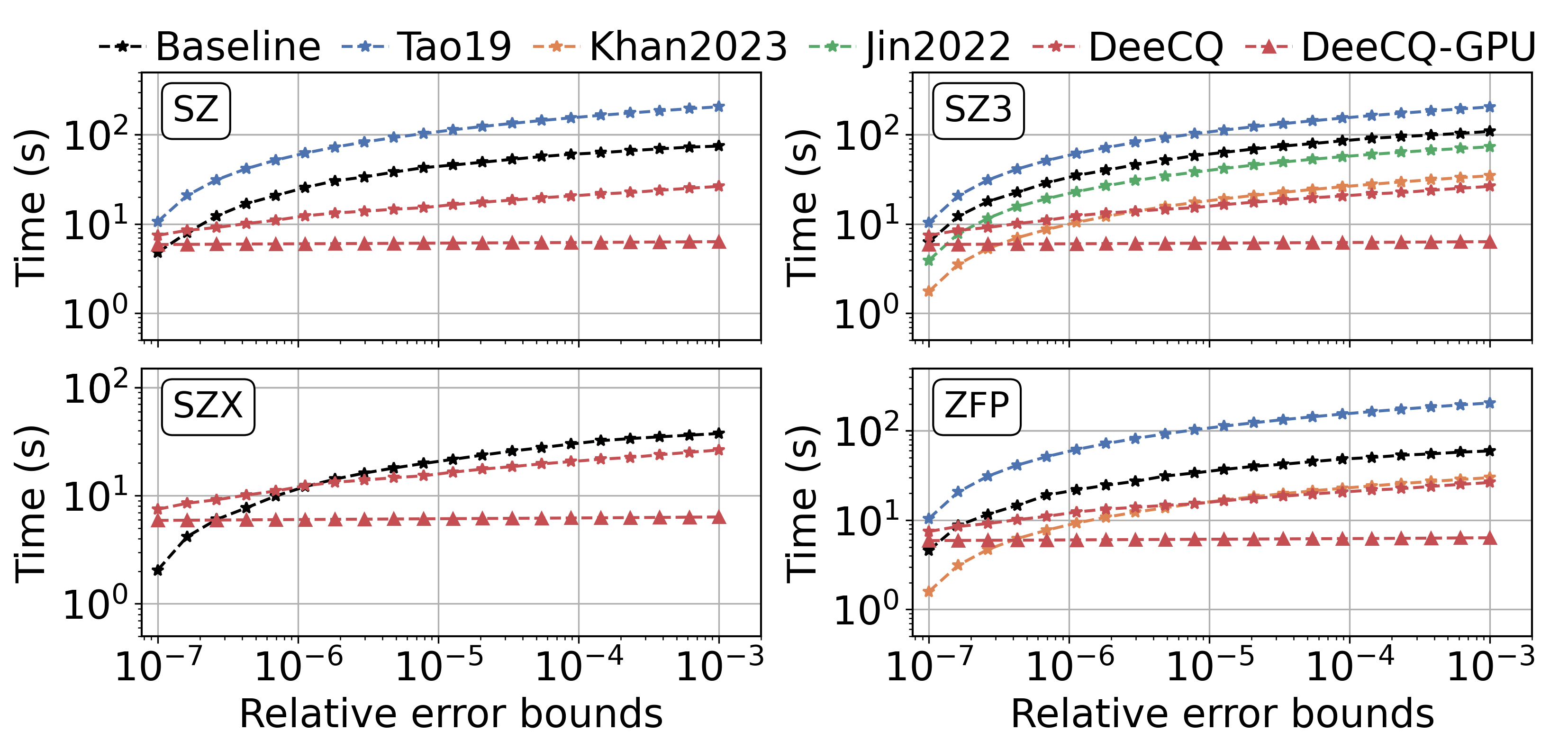}
    \vspace{-15pt}
    \caption{The cumulative time of CR prediction across 20 error bounds. Dataset: Dark matter density.}
    \vspace{-10pt}
    \label{fig:7}
\end{figure}

We also evaluate the cumulative prediction time for CR estimation across 20 relative error bounds using DeepCQ, Khan2023, and Jin2022 on SZ3 (Fig.~\ref{fig:7}). For reference, we include the compression time as a baseline. All timings are measured in wall-clock time, including data I/O. DeepCQ introduces an initial overhead of approximately 5 seconds to load the pre-trained model, but its inference time scales significantly more slowly with the number of error bounds compared to other methods.
Since our surrogate model is natively trainable and deployable on GPUs, we further report performance of GPU-based prediction, noted \texttt{DeepCQ\_GPU}. Although GPU execution incurs a slightly higher initialization cost due to CPU-GPU data transfer, the total prediction time grows much more slowly with the number of error bounds. This highlights DeepCQ’s superior scalability and efficiency---an advantage unmatched by prior methods.

\begin{mdframed}[linecolor=black, backgroundcolor=gray!2, roundcorner=6pt]
\textbf{\textit{Takeaway \#3}:} The proposed framework delivers highly efficient and scalable compression quality prediction across multiple error bounds and large-scale datasets, enabling rapid exploration of compressors and configurations within expansive search spaces.
\end{mdframed}

\subsection{Ablation Study} \label{sec:4.5}

\subsubsection{Training Overhead Reduction}

Table~\ref{tab:training} summarizes the training overhead of our proposed two-stage approach compared to the baseline, where the entire model is retrained from scratch for each configuration, using the Nyx application as an example. All models are trained for 150 epochs on 8 GPUs for a fair comparison. The baseline approach requires approximately 0.71 hours to train a single model for one data field, metric, and compressor. In contrast, the proposed two-stage approach first trains a shared feature extraction backbone (3.69 hours), followed by training of metric prediction heads, each requiring approximately 20 minutes (0.33 hours).

\begin{table}[h]
\centering
\setlength{\tabcolsep}{4pt}
\renewcommand{\arraystretch}{1.4}
\caption{The training overhead for Nyx (in hours).}
\vspace{-5pt}
\label{tab:training}
\begin{threeparttable}
\resizebox{.95\columnwidth}{!}{%
\begin{tabular}{ccccc}
\hline \hline
\multirow{2}{*}{\textbf{(\#F, \#C, \#M)}} & \multirow{2}{*}{\textbf{Baseline}} & \multicolumn{3}{c}{\textbf{Two-stage model training}} \\ \cline{3-5} 
                                          & & \textbf{Backbone} & \textbf{Prediction heads} & \textbf{Total} \\ \hline
(1, 1, 1)    & \underline{0.71}          & 1.15   & 0.33   & {1.48}                    \\ \hline
(1, 1, 3)    & \textit{\underline{2.13}} & 1.15   & 0.99   & \textit{\underline{2.14}} \\ \hline
(1, 5, 3)    & {10.7}                    & 1.15   & 4.95   & \underline{6.10}          \\ \hline
(4, 5, 5)    & {42.6}                    & 3.69   & 19.8   & \underline{23.49}         \\ \hline \hline
\end{tabular}%
}
\begin{tablenotes}
    \footnotesize
    \item \parbox{.8\textwidth}{(\#F, \#C, \#M) denote the numbers of {fields}, compressors, and metrics.}
    \item The method with lower overhead in each test case is underlined.
\end{tablenotes}
\end{threeparttable}
\end{table}

The two-stage design provides a substantial efficiency gain as the number of compressors and quality metrics increases, since the costly backbone training is performed only once. While the two methods perform similarly for small-scale configurations, the advantage of our approach grows rapidly with scale. For instance, across four data fields, five compressors, and three quality metrics, \textbf{the total training time is reduced from 42.6 hours to 23.49 hours, yielding a 1.81× speedup}.

\begin{mdframed}[linecolor=black, backgroundcolor=gray!2, roundcorner=6pt]
\textbf{\textit{Takeaway \#4}:} The proposed two-stage framework significantly reduces training overhead across multiple compressors, metrics, and datasets, making it well-suited for large-scale diverse prediction tasks in post-hoc analysis.
\end{mdframed}

\subsubsection{Improvement on Time-Evolving Data using MoE}

Table~\ref{tab:cr_comparison} compares the CR prediction error for the Nyx application, with and without incorporating the MoE architecture. Overall, the MoE-based model achieves consistently lower prediction errors than the baseline model using a simple MLP as the final prediction layer, demonstrating enhanced generalization across temporally evolving data. As discussed in Section~\ref{sec:3.5}, integrating MoE into the Pred-NN design increases representational capacity and enables the model to better capture variations in data distributions over time.

While the MoE-based model exhibits slightly higher errors in a few edge cases (e.g., CR prediction for SPERR on the Velocity x field), it maintains strong overall performance, achieving lower MAPE in the majority of scenarios. These results confirm the robustness of the MoE architecture under diverse data characteristics and temporal dynamics.

\begin{table}[h]
\setlength{\tabcolsep}{4pt}
\centering
\renewcommand{\arraystretch}{1.5}
\caption{CR prediction error with or without MoE.}
\vspace{-5pt}
\label{tab:cr_comparison}
\begin{threeparttable}
\resizebox{\columnwidth}{!}{%
\begin{tabular}{ccccccccccc}
\hline \hline
\multirow{2}{*}{\textbf{Fields}} 
& \multicolumn{2}{c}{\textbf{SZ}} 
& \multicolumn{2}{c}{\textbf{SZ3}} 
& \multicolumn{2}{c}{\textbf{SZX}} 
& \multicolumn{2}{c}{\textbf{ZFP}} 
& \multicolumn{2}{c}{\textbf{SPERR}} \\ \cline{2-11} 
     & B                & M                & B    & M                & B     & M                & B     & M                & B                & M                 \\ \hline
BD   & 8.17             & \underline{5.33} & 9.31 & \underline{5.82} & 11.20 & \underline{9.64} & 11.57 & \underline{9.45} & 13.11            & \underline{11.05} \\ \hline
DMD  & \underline{4.68} & 4.96             & 8.05 & \underline{3.54} & 5.98  & \underline{3.10} & 2.61  & \underline{5.07} & \underline{1.66} & 3.96              \\ \hline
VX   & \underline{1.31} & 1.38             & 3.73 & \underline{2.20} & 4.03  & \underline{1.89} & 3.15  & \underline{2.84} & \underline{2.72} & 4.04              \\ \hline
TEMP & 5.45             & \underline{3.54} & 3.18 & \underline{3.13} & 3.23  & \underline{2.34} & 5.54  & \underline{4.51} & 4.41             & \underline{1.51}  \\ \hline \hline
\end{tabular}%
}
\begin{tablenotes}
    \footnotesize
    \item BD, DMD, VX, TEMP represent the data fields from Nyx application.
    \item \textit{B} and \textit{M} represent the Pred-NN with \textit{simple MLP} and \textit{MoE}, respectively.
    \item The better-performing method in each test case is underlined.
\end{tablenotes}
\end{threeparttable}
\vspace{-10pt}
\end{table}

\begin{mdframed}[linecolor=black, backgroundcolor=gray!2, roundcorner=6pt]
\textbf{\textit{Takeaway \#5:}} Incorporating the MoE architecture improves model generalization across temporally evolving datasets by enabling specialized feature learning and adaptive prediction, achieving lower overall error across most compressors and data fields.
\end{mdframed}

\subsubsection{Generalization Across Data Fields}
We next evaluate the generalization capability of our framework for predicting compression quality metrics across data fields within each application. Specifically, we train a shared metric prediction model using training data aggregated from all fields in a given application and report prediction errors for four representative fields from Nyx and Hurricane in Table~\ref{tab:allfields}.

Since compression quality strongly depends on the intrinsic patterns of the input data, cross-field training can introduce bias in prediction. Nevertheless, the results show that predictions for PSNR and SSIM remain generally accurate, especially for fields whose compression behavior is strongly correlated with the error bound. These findings demonstrate the feasibility of employing shared metric prediction models across multiple fields within a single application, thereby further reducing model training and management overhead.

However, noticeable variations across fields highlight an ongoing challenge---differences in spatial and statistical patterns can still degrade prediction accuracy for certain metrics and compressors. This underscores the need for further exploration of architectures and training strategies that enhance cross-field robustness and feature transferability.

\begin{table}[t]
\centering
\setlength{\tabcolsep}{4pt}
\renewcommand{\arraystretch}{1}
\caption{Prediction error (\%) of generalization across all fields.}
\vspace{-5pt}
\label{tab:allfields}
\begin{threeparttable}
\resizebox{\columnwidth}{!}{%
\begin{tabular}{ccc cc cc}
\hline \hline
  \multirow{2}{*}{\textbf{Fields}}
& \multicolumn{2}{c}{\textbf{SPERR}} 
& \multicolumn{2}{c}{\textbf{ZFP}} 
& \multicolumn{2}{c}{\textbf{SZ3}} 
\\
\cline{2-3} \cline{4-5} \cline{6-7}
& PSNR & SSIM 
& PSNR & SSIM 
& PSNR & SSIM 
\\
\hline
Baryon density         & 11.20 & 31.78    & 3.62 & 3.02      & 4.49 & 119.99 \\ 
Dark matter density    & 0.37  & 2.15     & 1.74 & 3.95      & 2.10 & 28.74  \\ 
Velocity x             & 2.20  & 0.28     & 1.50 & 0.59      & 1.87 & 0.67   \\ 
Temperature            & 0.83  & 9.33     & 1.49 & 0.47      & 1.56 & 2.54   \\ 
QICE                   & 3.64  & 56.76    & 2.04 & 32.38     & 2.85 & 45.83  \\ 
QSNOW                  & 1.84  & 12.43    & 2.14 & 19.00     & 3.12 & 12.45  \\ 
U                      & 1.21  & 0.31     & 1.63 & 0.34      & 0.46 & 0.66   \\ 
TC                     & 0.40  & 0.70     & 1.56 & 1.00      & 0.55 & 1.14   \\ 
\hline \hline
\end{tabular}%
}
\end{threeparttable}
\vspace{-10pt}
\end{table}

\begin{mdframed}[linecolor=black, backgroundcolor=gray!2, roundcorner=6pt, innermargin =-5cm]
\textbf{\textit{Takeaway \#6}:} Surrogate-based metric prediction holds the potential to be shared across data fields, given that data from different fields react to error bounds with similar sensitivity. Further study is needed for accurate and robust prediction across data fields.
\end{mdframed}

\section{Limitations and Future Work}\label{sec:discussion}

\noindent \textbf{\textit{Generalization Across Timesteps and Data Fields.}}
While the proposed MoE-based prediction model improves robustness against moderate temporal variations, its ability to generalize across significantly different timesteps or heterogeneous data fields remains limited. Future work will focus on systematically analyzing these variations to better understand their impact on compression quality, and enhance model generality across time and physical variables.

\noindent \textbf{\textit{Compressor-Specific Metric Prediction.}}
Certain compressors (e.g. SZx and ZFP) exhibit non-smooth quality–error relationships due to intrinsic compression algorithms. A general-purpose prediction head can not provide adequate capability to capture the unique behaviors of such compressors. Developing compressor-specific prediction architectures that explicitly account for such non-linear or discontinuous behaviors is a promising direction for improving predictive fidelity.

\noindent \textbf{\textit{System-Level Support and Continuous Validation.}}
The prediction accuracy of surrogate models fundamentally depends on the diversity and quality of training data. Currently, the framework provides limited mechanisms for continuous validation or detection of data drift during long-term operation. Future extensions should integrate system-level support, such as automated monitoring and online retraining, within HPC workflows and scientific simulation pipelines to maintain reliability and adapt to evolving data characteristics. 

\section{Conclusion} \label{sec:conclusion}

In this work, we present DeepCQ, a deep surrogate-based framework for accurate and efficient prediction of compression quality across compressors, metrics, and scientific datasets. By decoupling feature extraction from metric prediction through a two-stage design, DeepCQ significantly reduces training overhead and enables modular, scalable inference.
Evaluation results demonstrate its outstanding prediction accuracy and efficiency. The application-specific backbone further supports lightweight, customizable models for diverse downstream tasks and can serve as a foundation for domain-specific extensions, fostering data-driven scientific discovery through knowledge transfer.

\bibliographystyle{IEEEtran}
\bibliography{main}

\end{document}